\pdfoutput=1
\PassOptionsToPackage{table}{xcolor}

\documentclass[11pt]{article}

\usepackage[final]{acl}

\usepackage{times}
\usepackage{latexsym}

\usepackage[T1]{fontenc}

\usepackage[utf8]{inputenc}

\usepackage{microtype}

\usepackage{inconsolata}

\usepackage{graphicx}

\usepackage{multirow}
\usepackage{tcolorbox}
\usepackage[table]{xcolor}
\usepackage{subcaption}
\usepackage{amsmath}
\usepackage{booktabs}

\title{Masking in Multi-hop QA: \\ An Analysis of How Language Models Perform with Context Permutation}

\author{
  Wenyu Huang$^{1}$,
  Pavlos Vougiouklis$^{2}$, 
  Mirella Lapata$^{1}$, 
  Jeff Z. Pan$^{1,2}$ \\
  $^1$School of Informatics, University of Edinburgh \\ 
  $^2$Huawei Edinburgh Research Centre, Poisson Lab, CSI, UK
  \\\texttt{w.huang@ed.ac.uk}, \texttt{pavlos.vougiouklis@huawei.com}, \texttt{mlap@inf.ed.ac.uk}, \\\texttt{http://knowledge-representation.org/j.z.pan/}
}

\begin{document}
\maketitle
\begin{abstract}
Multi-hop Question Answering (MHQA) adds layers of complexity to question answering, making it more challenging. When Language Models (LMs) are prompted with multiple search results, they are tasked not only with retrieving relevant information but also employing multi-hop reasoning across the information sources. Although LMs perform well on traditional question-answering tasks, the causal mask can hinder their capacity to reason across complex contexts. In this paper, we explore how LMs respond to multi-hop questions by permuting search results (retrieved documents) under various configurations. Our study reveals interesting findings as follows: 1) Encoder-decoder models, such as the ones in the Flan-T5 family, generally outperform causal decoder-only LMs in MHQA tasks, despite being significantly smaller in size; 2) altering the order of gold documents reveals distinct trends in both Flan T5 models and fine-tuned decoder-only models, with optimal performance observed when the document order aligns with the reasoning chain order;
3) enhancing causal decoder-only models with bi-directional attention by modifying the causal mask can effectively boost their end performance. In addition to the above, we conduct a thorough investigation of the distribution of LM attention weights in the context of MHQA. Our experiments reveal that attention weights tend to peak at higher values when the resulting answer is correct. We leverage this finding to heuristically improve LMs' performance on this task. Our code is publicly available at \url{https://github.com/hwy9855/MultiHopQA-Reasoning}.
\end{abstract}

\section{Introduction}
\label{sec:intro}

\begin{figure}
    \centering    
    \begin{subfigure}[b]{0.23\textwidth} 
    \centering 
    \includegraphics[width=\linewidth]{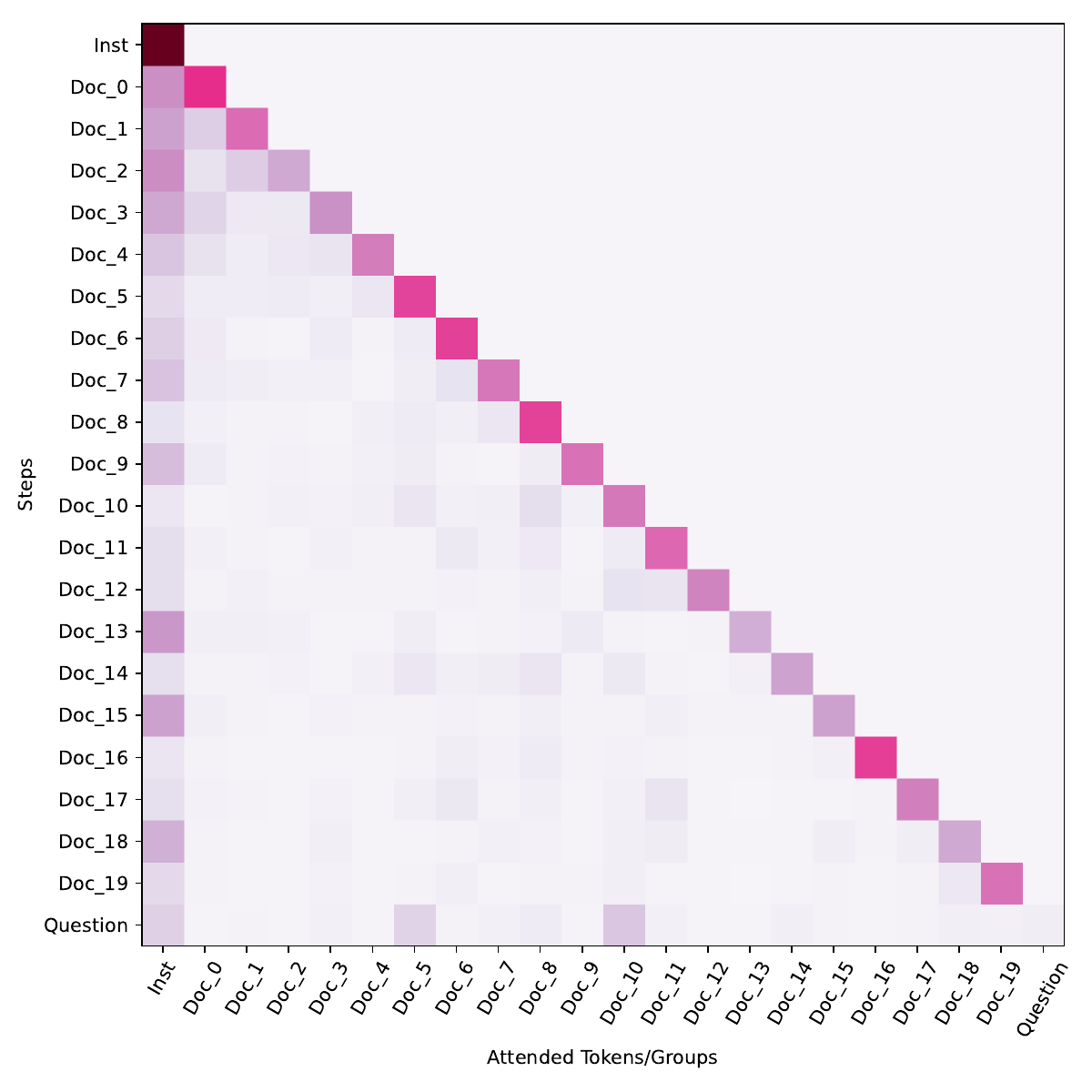}
    \caption{Qwen 2.5, head 3}
    \end{subfigure} 
    \hfill
    \begin{subfigure}[b]{0.23\textwidth} 
    \centering 
    \includegraphics[width=\linewidth]{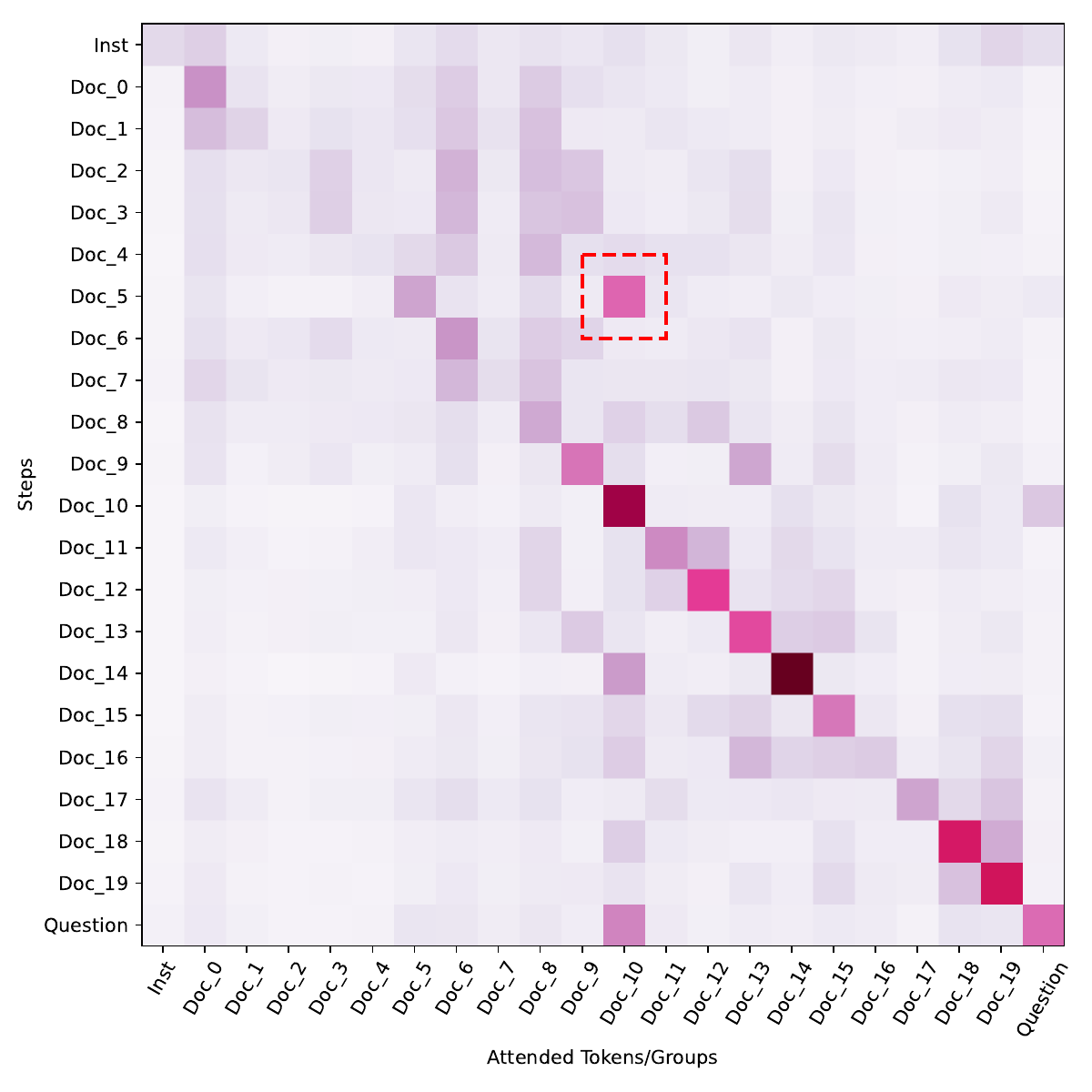}
    \caption{FlanT5, head 1}
    \end{subfigure} 
    \caption{Context attention distribution of Qwen 2.5 1.5B and FlanT5 large for the same question, both captured from the last layer (last encoder layer for FlanT5). The gold documents are Doc\_5 and Doc\_10, where the reasoning direction is Doc\_10 $\rightarrow$ Doc\_5. With bi-directional attention, FlanT5 allows Doc\_5 to assign attention and ``see'' Doc\_10 (red dashed box), whereas Qwen 2.5 with causal mask cannot.}
    \label{fig:intro}
\end{figure}

Language Models (LMs) based on Transformer architectures \citep{NIPS2017_3f5ee243} have become essential tools for a wide variety of tasks, including question answering and conversational search \citep{NEURIPS2023_9cb2a749, yi2024surveyrecentadvancesllmbased, mo2024surveyconversationalsearch}. Among the different architectures, causal decoder-only configurations have become a popular choice for many of the most widely known LM families \citep{llama3, qwen}. 

LMs exhibit a strong ability to reason within their input context, allowing for adaptability and effective generalisation across a wide range of tasks \cite{OpenAI2024}. Albeit these capabilities, previous studies have highlighted significant challenges regarding \textit{the extent to which LMs can reason across different input contexts} \citep{zhang2023sirenssongaiocean, kaddour2023challengesapplicationslargelanguage}.
An exemplary case is the ``lost in the middle'' problem \citep{liu-etal-2024-lost},
where crucial information positioned in the middle of the context may be overlooked by LMs. As LMs are increasingly utilised in complex scenarios, their ability to reason across different contexts becomes critically important. The general Retrieval-augmented Generation (RAG) framework \cite{Lewis2020,Li2024} has recently become the cornerstone of many search-based conversational agents, such as Copilot and Doubao. In this setting, LMs frequently need to synthesise information from multiple retrieved search results to provide coherent answers.

This study conducts an in-depth analysis of how LMs reason across various contexts, specifically focusing on multi-hop question answering (MHQA). Within the RAG framework, MHQA necessitates synthesising knowledge from multiple documents, presenting a more complex level of information integration than other more trivial question-answering tasks. To successfully perform MHQA, LMs must not only identify the most relevant documents within a given context but also reason with the information from these documents to determine the correct answer.

A critical research question arises from the architecture of modern causal decoder-only Transformer-based LMs, which use a causal mask during both training and inference. This constraint hinders these models from performing bi-directional encoding \citep{DBLP:journals/jmlr/RaffelSRLNMZLL20}, as opposed to traditional encoder-decoder architectures that can capture interactions between documents more effectively (cf. Figure~\ref{fig:intro}). This leads us to examine whether these widely used causal decoder-only LMs are limited by this constraint. Further, what might be the impact if we replaced the causal mask with a bi-directional (i.e. prefix) mask?

To address these inquiries, we conduct a comprehensive investigation about how LMs model the MHQA task. Initially, we evaluate the MHQA performance of three widely adopted open-source LM families: the Flan-T5 \citep{DBLP:journals/jmlr/ChungHLZTFL00BW24} family, representing the traditional encoder-decoder architecture, and the Qwen2.5 \citep{qwen} and Llama 3.x \citep{llama3} families, exemplifying two popular causal decoder-only LMs. We design three types of document permutations to investigate the LMs' behaviour in terms of: 1) the order of gold documents\footnote{In the context of MHQA, they usually refer to the documents needed to answer the single-hop questions into which the original multi-hop question is decomposed.}, 2) the distance between them, and 3) their completeness. Our observations include the following: 1) The encoder-decoder model (Flan T5) is a superior MHQA solver when no fine-tuning takes place; 2) fine-tuned LMs tend to favour forward-placed documents \citep{10.5555/3692070.3692325} (i.e., when the order of gold documents in the context mirrors the order of the reasoning chain, cf. Figure~\ref{fig:exp_design}), a trend also observed in the Flan T5 models; 3) bi-directional attention with prefix mask can benefit LMs in MHQA tasks and offers better robustness when the order of gold documents is altered; 4) the distance between gold documents significantly affects performance, and;
5) while the removal of the first hop document reduces MHQA performance, a relatively high level of correctness is still maintained.

Building on the above observations, we delve deeper to analyse how LMs perform MHQA by examining the attention distribution across layers. Our findings reveal that LMs typically assign higher attention score weights to at least one document when they correctly answer a multi-hop question. By sampling answers with different input document permutations, and retaining the answers for the inputs for which the LM assigned the largest peak attention score, we increased the accuracy of Qwen 7B from 28.6\% to 33.7\%. 

\section{Related Works}
\subsection{RAG and MHQA}
The RAG framework has been widely used to inject external knowledge into LMs \citep{gao2024retrievalaugmentedgenerationlargelanguage}, and mitigate their hallucination tendencies \citep{10.1145/3703155}. In this framework, LMs generate a response by conditioning it on the top search results, which are provided as input context. This setup has shown promising performance in a variety of tasks, including knowledge graph question answering \citep{luo2024reasoning, huang-etal-2024-less}, open-domain dialogue generation \citep{wang2024unimsragunifiedmultisourceretrievalaugmented}, and multi-hop question answering \citep{trivedi-etal-2022-musique-fixed}. Some studies have focused on investigating how LMs use the input context, identifying discrepancies in how different parts of the input are processed. For instance, \citet{mallen-etal-2023-trust} find that retrieval may sometimes harm model predictions and \citet{liu-etal-2024-lost} report that LMs suffer from a ``lost in the middle'' predicament. However, these works focus on simpler, one-hop question-answering tasks that do not require reasoning across distant contextual hops. Going a step further, \citeauthor{10.5555/3692070.3692325} demonstrate that the order of the premise matters when logical reasoning is expected by LMs, while \citet{baker2024lostmiddleinbetweenenhancing} find that LMs also ``lost in between'' when the distance between information is increased.
In this work, we build upon existing methods to explore how language models reason over their input context to address multi-hop questions.

\subsection{Drawbacks of Causal Language Model}
A causal language model is equipped with a causal mask that prevents tokens in the context to \textit{see} future context \citep{DBLP:journals/jmlr/RaffelSRLNMZLL20}. This design harms the performance on complex tasks that require rich contextualized representations \citep{li-etal-2023-chatgpt, qorib-etal-2024-decoder}. A common way to mitigate this issue is to prompt the model by repeating the context in the input. \citeauthor{xu-etal-2024-reading} show that repeating the context improves the LMs ability in reasoning. \citeauthor{springer2024repetitionimproveslanguagemodel} prove that repeating the context can also improve embedding quality. To sidestep the unnecessary cost associated with repeating the input context, researchers have started to introduce bi-directional attention into decoder-only models. \citeauthor{behnamghader2024llmvec} and \citeauthor{muennighoff2024generativerepresentationalinstructiontuning} successfully apply bidirectional attention in decoder-only models to generate text embeddings, and observe better quality in a variety of tasks. In this work, we explore how bi-directional attention can enhance language models' performance in the MHQA task, uncovering valuable heuristics that could improve model outcomes.

\section{Preliminary}
\subsection{Multihop Question Answering}
We formally define the MHQA task $\mathcal{T}$. The input of $\mathcal{T}$ has two parts, a question $q$, and $n$ documents $\mathcal{D}=\left \{d_1, \ldots, d_{n} \right \}$, some of which are relevant to $q$ and some are not. For answering $q$, the information from $m$ documents, s.t. $m < n$, is mandatory.
We accomplish $\mathcal{T}$ by prompting LMs with the concatenation of $q$ and $\mathcal{D}$ in the input context.

\subsection{Grouped Attention Weight}
For better investigating which documents are more \textit{heavily attended} by LMs, we compute grouped attention weights between token blocks. There are several blocks, including the instruction block, document blocks, question block, and prediction blocks. Please note that for prediction blocks, we directly use the prediction tokens that are not grouped, since the prediction blocks may contain other tokens besides the answer tokens. For attention between block $X$ and block $Y$, the grouped attention weight is computed as:
\begin{align}
    \text{GA}_{l,h}(X, Y) = \frac{1}{|X|}\sum_{t_X\in X}\sum_{t_Y\in Y}\text{Attention}_{l,h}(t_X, t_Y)
    \label{eq:ga}
\end{align}where $l$, $h$ denote the decoder layer and the attention head, $|X|$ is used to normalize grouped attention values, $t_X$ and $t_Y$ are tokens of block $X$ and $Y$. By grouping attention with~(\ref{eq:ga}), we make sure:
\begin{align}
    \sum_Y\text{GA}_{l,h}(X, Y) = 1
\end{align}
Using Eq.~(\ref{eq:ga}), we can 
understand which context part contributes more to the next token prediction. 

\subsection{Information Contribution Score}
To investigate how much information from each document is captured during the MHQA task, we introduce the Information Contribution (IC) score based on the grouped attention scores.

For layer $l$, document token group $d$, the Information Contribution (IC) score is by defined:
\begin{align}
    \text{IC}_{l}(d) = \frac{1}{|A||H|}\sum_{h\in H}\sum_{a\in A}\text{GA}_{l,h}(a, d)
    \label{equ:ic}
\end{align}
where $H$ is the set of attention heads and $A$ is the set of answer tokens in the prediction.

\section{Experimental Setup}

We conduct experiments using both encoder-decoder LMs (from the Flan-T5 family) and causal decoder-only LMs (Qwen 2.5 family and Llama 3.x family). For the Qwen 2.5 family, we select to use five LMs with sizes that range from 0.5B to 14B. For the Llama 3.x family, we use 1B and 3B from Llama 3.2 and 8B from Llama 3.1. The specifications of these models are included in Table~\ref{tab:llm_spec}. We use the instruction-tuned variants of all the models included in this study. We investigate 4 setups for inference:

\paragraph{Answer Only}The model is forced to directly generate the answer in the following format: \texttt{\textbackslash box\{$\langle$answer$\rangle$\}}.
\paragraph{CoT}Zero-shot Chain of Thought prompting is used to ask the model to first generate reasoning steps and then provide the final answer in the same format: \texttt{\textbackslash box\{$\langle$answer$\rangle$\}}.
\paragraph{Finetuned}We use the MuSiQue training set to train the models.
\paragraph{Finetuned + Bi}We replace the original causal mask of the model with bi-directional attention facilitated by a 2D mask, as follows:\begin{align}
    M_{i,j} = 
    \begin{cases}
    0 & \text{if } i \geq j \\ 
    0 & \text{if } i \leq c, j \leq c \\ 
    1 & \text{otherwise} 
    \end{cases}
\end{align}where $c$ is the context length. With the new mask, the model is then converted to a prefix LM \citep{DBLP:journals/jmlr/RaffelSRLNMZLL20}
The model is trained with the same data as in the \textbf{Finetuned} setup.

For all finetuning experiments, we use LoRA with $r=8$ and $\alpha=16$ and train for 5 epochs with a learning rate $2e-5$ and a batch size 1.

\subsection{Dataset}
We instantiate the MHQA task with the MuSiQue \citep{trivedi-etal-2022-musique-fixed} dataset, which contains multi-hop questions from 2-hop to 4-hop. In the original dataset, for each question, $2-4$ gold documents are provided, each containing evidence for each decomposed question (hop). Additionally, distractor documents are included to add noise, forming a context with up to 20 documents in total, presented in no specific order.
We use the answerable set in all the experiments, and keep the data split unchanged in terms of the training and development set.
For the finetuning experiments, we use the original training set with 19,938 queries. For all the experiments, we report performance on the development set which consists of 2,417 queries.

\subsection{Metrics}
We use the accuracy metric (\textbf{Acc}) to measure MHQA performance. For the \textbf{Answer Only} setup, we treat the answer inside \texttt{\textbackslash boxed\{$\langle$answer$\rangle$\}} as the prediction. For \textbf{CoT} setup, we treat the last \texttt{\textbackslash boxed\{$\langle$answer$\rangle$\}} as the prediction. Please note that some models do not follow the CoT instruction well\footnote{Small Qwen models (0.5B and 1.5B) generally do not apply CoT reasoning. Llama models do not always follow the instruction to place the final answer in \texttt{\textbackslash boxed\{$\langle$answer$\rangle$\}}.}. In such cases, we compute \textbf{Acc} by finding if the reference answer is included in the last line of the prediction. For finetuning-based models and encoder-decoder models, we compute the exact match as the accuracy of the prediction.

\section{Does Permutation Change LM's Mind?}
\begin{figure}
    \centering
    \includegraphics[width=\linewidth]{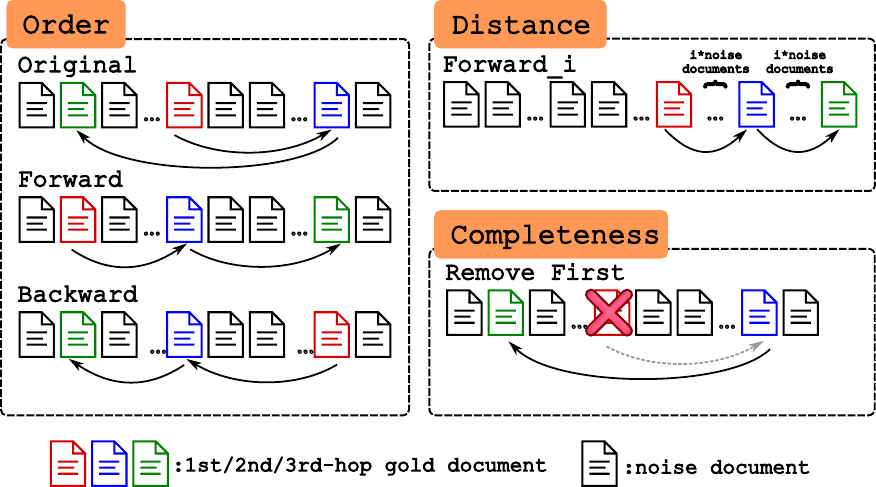}
    \caption{Context permutation design. Take 3-hop question for example. Forward setup follows the reasoning chain while Backward setup is the reverse of Forward setup. Forward\_i control the distance between gold documents, and Remove First removing the document that support first hop question.}
    \label{fig:exp_design}
\end{figure}
We designed three distinct context permutation settings to assess LMs' abilities to comprehend different aspects of the context, as shown in Figure~\ref{fig:exp_design}.
\paragraph{Order of gold documents}
In an idealised scenario, we would expect the gold documents to be provided in the same order as the reasoning hops. Our hypothesis is that the effect of the causal mask will be minimised by providing the gold documents with the reasoning chain order. We refer to this setting as \textbf{Forward}. In contrast to this setting, we test a setting in which the documents are placed in the reverse order of Forward, hypothesising that the counter-intuitive reasoning consistency should make the task more challenging for the involved LMs. The reversed setting is called \textbf{Backward}. Finally, we have a setting where we keep the order of the gold documents as in the original dataset, to which we refer as \textbf{Original}.

\paragraph{Distance of gold documents}
Besides the order, the distance between the input gold documents is not guaranteed in real-world MHQA tasks. To investigate how the distance between gold documents affects the LM, we design a series of \textbf{Forward}\_$i$ settings, where we fix the order of documents to be Forward, and, subsequently, ensure that the final hop document is placed at the end of the context. After that, between each gold document, we inject $i$ noise documents (i.e. the set of noisy documents remain the same as in the previous settings). We select $i=\{0, 1, 2, 3, 4, 5\}$, where $i=0$ stands for no noise between gold documents. 

\paragraph{Completeness of gold documents}
To better understand how LMs answer multi-hop questions, we want to know if they are really doing multi-hop reasoning, or just \textit{guess} an answer. To evaluate this, we design a setting in which the knowledge in the input context is incomplete in terms of answering the original question: \textbf{Remove First}, which removes the first hop document.\footnote{Please note that, in a formal definition, removing an element does not qualify as context permutation. We use the term ``permutation'' here for convenience in this context.}

\begin{table*}[t]
    \centering
    \small
    \begin{tabular}{l|ccc|ccc|ccc|ccc}
    \toprule
        \multirow{2}{*}{\textbf{Model}} & \multicolumn{3}{c|}{\textbf{Answer Only}} &  \multicolumn{3}{c|}{\textbf{CoT}} & \multicolumn{3}{c|}{\textbf{Finetuned}} & \multicolumn{3}{c}{\textbf{Finetuned + Bi}} \\
         & $\Delta_{B}$ & Acc & $\Delta_{F}$ 
         & $\Delta_{B}$ & Acc & $\Delta_{F}$
         & $\Delta_{B}$ & Acc & $\Delta_{F}$
         & $\Delta_{B}$ & Acc & $\Delta_{F}$\\
        \midrule
        Qwen2.5 0.5B & \cellcolor{green!2}{0.21} & 8.94 & \cellcolor{red!5}{-0.58} & \cellcolor{red!2}{-0.21} & 12.91 & \cellcolor{red!7}{-0.79} & \cellcolor{red!43}{-4.34} & 27.14 & \cellcolor{green!37}{3.72} & \cellcolor{red!9}{-0.91} & 30.30 & \cellcolor{green!4}{0.41} \\
        Qwen2.5 1.5B & \cellcolor{red!10}{-1.08} & 20.36 & \cellcolor{red!7}{-0.70} & 0.0 & 22.76 & \cellcolor{red!10}{-1.03} & \cellcolor{red!26}{-2.61} & 44.06 & \cellcolor{green!18}{1.86} & \cellcolor{green!0}{0.04} & 44.78 & \cellcolor{green!12}{1.20} \\
        Qwen2.5 3B & \cellcolor{red!8}{-0.87} & 19.78 & \cellcolor{green!7}{0.74} & \cellcolor{red!20}{-2.03} & 24.82 & \cellcolor{red!4}{-0.46} & \cellcolor{red!13}{-1.37} & 50.23 & \cellcolor{green!29}{2.98} & \cellcolor{red!17}{-1.74} & 52.15 & \cellcolor{green!16}{1.70} \\
        Qwen2.5 7B & \cellcolor{red!17}{-1.78} & 28.59 & \cellcolor{green!7}{0.74} & \cellcolor{red!20}{-2.03} & 36.24 & \cellcolor{green!20}{2.03} & \cellcolor{red!23}{-2.36} & 58.05 & \cellcolor{green!14}{1.41} & \cellcolor{red!14}{-1.45} & 62.96 & \cellcolor{green!3}{0.33} \\
        Qwen2.5 14B & \cellcolor{green!1}{0.12} & 37.07 & \cellcolor{red!2}{-0.29} & \cellcolor{green!10}{1.08} & 39.22 & \cellcolor{green!6}{0.62} & \cellcolor{red!16}{-1.65} & 64.34 & \cellcolor{green!10}{1.03} & \cellcolor{green!2}{0.29} & 64.88 & \cellcolor{green!0}{0.08} \\
        \midrule
        Llama3.2 1B & \cellcolor{green!8}{0.83} & 11.21 & \cellcolor{red!8}{-0.83} & \cellcolor{red!0}{-0.04} & 11.96 & \cellcolor{red!2}{-0.21} & \cellcolor{red!19}{-1.99} & 33.06 & \cellcolor{green!17}{1.74} & \cellcolor{red!4}{-0.41} & 40.85 & \cellcolor{green!6}{0.62} \\
        Llama3.2 3B & \cellcolor{red!16}{-1.61} & 25.73 & \cellcolor{green!7}{0.79} & \cellcolor{red!6}{-0.62} & 31.65 & \cellcolor{green!11}{1.12} & \cellcolor{red!19}{-1.99} & 54.57 & \cellcolor{green!16}{1.70} & \cellcolor{red!9}{-0.91} & 59.60 & \cellcolor{green!5}{0.58} \\
        Llama3.1 8B & \cellcolor{green!5}{0.54} & 36.37 & \cellcolor{red!9}{-0.95} & \cellcolor{red!6}{-0.62} & 44.60 & \cellcolor{red!2}{-0.25} & \cellcolor{red!21}{-2.11} & 63.51 & \cellcolor{green!12}{1.24} & \cellcolor{red!12}{-1.20} & 65.48 & \cellcolor{green!14}{1.41} \\ 
        \bottomrule
    \end{tabular}
    \caption{Overall MHQA performance on the MuSiQue development set. $\Delta_B$ and $\Delta_F$ are performance differences between original documents and re-ordered backward and forward documents respectively. {\color{green}{Green cells}} indicate performance improvement while {\color{red}{red cells}} indicate performance drop.}
    \label{tab:overall_em}
\end{table*}

\subsection{Order of Documents Matters}
\label{sec:order}
Table~\ref{tab:overall_em} shows performance across different setups. 

\paragraph{Non-finetuned}
Generally, the \textbf{Answer Only (AO)} results in the worst performance. In this setup, changing the order of the supporting documents leads to unstable performance differences, where forward-placed gold documents do not offer overall performance improvement consistently. 
With CoT prompting, most models get performance improvements. We note that Qwen2.5 0.5B, 1.5B and Llama3.2 1B do not follow the CoT instruction well, as the inclusion of the CoT instruction does not often lead to changes in the models' output.
Overall, decoder-only models with zero-shot CoT perform similarly to when they are directly prompted, and the order of documents appears to be invariant to their MHQA performance.

\paragraph{Finetuned (FT)}

By finetuning the models on the MuSiQue training set, the performance of all models improves. 
Interestingly, finetuned variants seem to benefit from the forward gold document setting, even though the training data are provided in the original order of the training split. To ensure that the order of the documents in the training set is not sequentially correlated with the forward (and backward) setting, we compute the average Spearman's rank correlation and Kendall's $\tau$ coefficients: $0.0013$ ($-0.0013$) and $0.0016$ ($-0.0016$) respectively, indicating that the benefits of the forward setup are \textit{emergent} through finetuning.

\paragraph{Finetuned + Bi (FT+Bi)}
When finetuning the models modified with bi-directional attention, we observe further performance improvements. Moreover, the models are more robust to document permutations, showing less variation in performance across the three inference setups.

\paragraph{Encoder-Decoder}
\begin{table}[t]
    \centering
    \small
    \begin{tabular}{l|ccc|l}
    \toprule
        \textbf{Model} & $\Delta_{B}$ & Acc & $\Delta_{F}$ & Qwen2.5 Acc \\
        \midrule

        FT5 small/80M & \cellcolor{red!19}{-1.94} & 20.11 & \cellcolor{green!18}{1.86} & 8.94 (0.5B) \\
        FT5 base/250M & \cellcolor{red!16}{-1.65} & 28.09 & \cellcolor{green!19}{1.90} & 20.36 (1.5B)  \\
        FT5 large/0.8B & \cellcolor{green!2}{0.25} & 40.01 & \cellcolor{green!3}{0.37} & 19.78 (3B) \\
        FT5 xl/3B & \cellcolor{red!24}{-2.44} & 47.33 & \cellcolor{green!21}{2.19} & 28.59 (7B) \\
        FT5 xxl/11B & \cellcolor{red!20}{-2.03} & 56.43 & \cellcolor{green!16}{1.65} & 37.07 (14B) \\
        
        \bottomrule
    \end{tabular}
    \caption{MHQA performance on the MuSiQue development set using Flan  
T5 models. 
    Qwen2.5 Acc is the Acc score from the Qwen2.5 family for reference. 
    }
    \label{tab:encdec_em}
\end{table}

\begin{figure}[t]
    \centering
    \includegraphics[width=\linewidth]{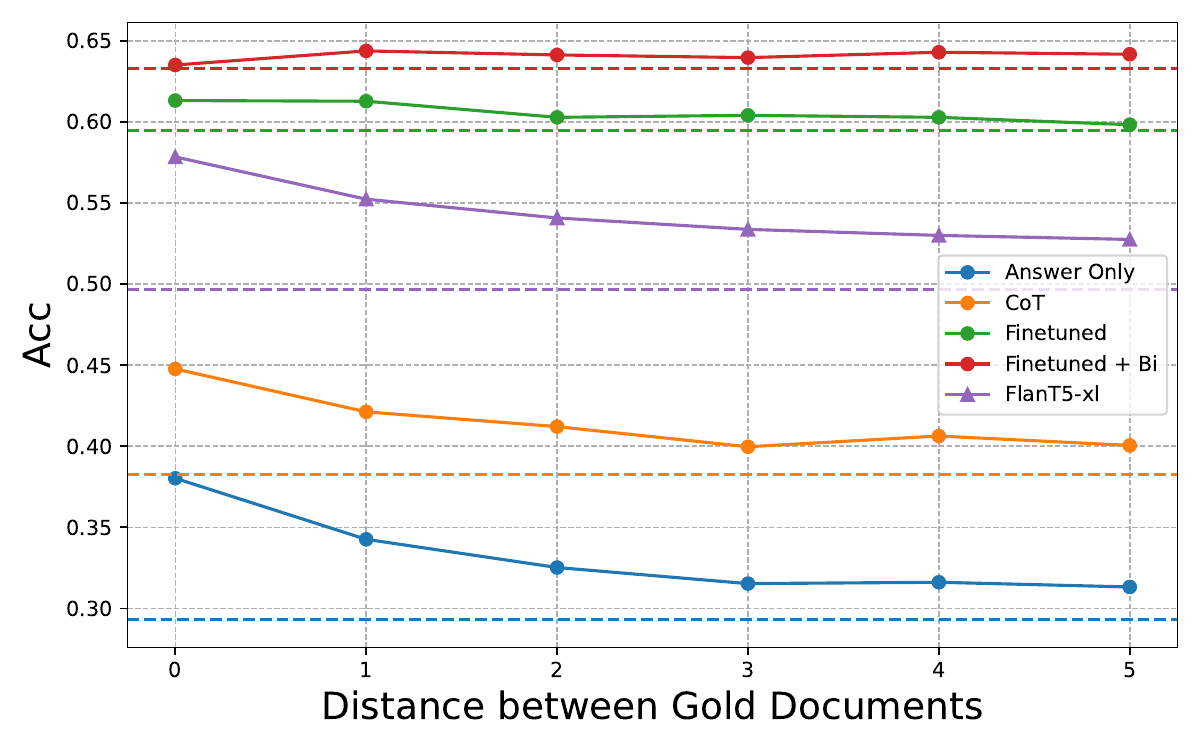}
    \caption{Acc of Qwen2.5 7B across different gold document settings. Each dashed line refers the Forward result of the setup with the same colour.
    The performance of non-finetuned models generally drops as the distance increases, while finetuned models show better robustness to the increase of distance.
    }
    \label{fig:distance}
\end{figure}
\begin{table}[t]
    \centering
    \small
    \begin{tabular}{l|ccc}
        \toprule
        \multirow{2}{*}{\textbf{Setup}} & \multicolumn{3}{c}{\textbf{Accuracy}} \\
        & 2-Hop & 3-Hop & 4-Hop \\ \midrule
         \multicolumn{4}{c}{\textit{w/o  first hop information in parametric knowledge}} \\ \midrule
        AO & \cellcolor{red!5}{26.0$\rightarrow$20.8} & \cellcolor{green!0}{28.6$\rightarrow$29.1} & \cellcolor{red!0}{29.7$\rightarrow$28.8} \\
        CoT & \cellcolor{red!25}{40.3$\rightarrow$14.8} & \cellcolor{red!11}{38.5$\rightarrow$27.1} & \cellcolor{green!0}{24.3$\rightarrow$24.3} \\
        FT & \cellcolor{red!27}{63.9$\rightarrow$36.2} & \cellcolor{red!3}{57.7$\rightarrow$54.4} & \cellcolor{green!0}{56.2$\rightarrow$57.2} \\
        FT+Bi & \cellcolor{red!33}{70.9$\rightarrow$37.7} & \cellcolor{red!1}{57.0$\rightarrow$55.7} & \cellcolor{red!3}{64.9$\rightarrow$61.7} \\
        FT5 xl & \cellcolor{red!14}{53.8$\rightarrow$39.1} & \cellcolor{red!1}{41.3$\rightarrow$39.7} & \cellcolor{green!0}{40.1$\rightarrow$40.1} \\
         \midrule
         \multicolumn{4}{c}{\textit{w/ first hop information in parametric knowledge}} \\ \midrule
        AO & \cellcolor{red!0}{33.6$\rightarrow$33.3} & \cellcolor{red!1}{30.8$\rightarrow$29.0} & \cellcolor{green!0}{22.8$\rightarrow$22.8} \\
        CoT & \cellcolor{red!10}{40.2$\rightarrow$30.0} & \cellcolor{green!8}{30.8$\rightarrow$39.7} & \cellcolor{green!3}{20.7$\rightarrow$23.9} \\
        FT & \cellcolor{red!5}{57.5$\rightarrow$52.2} & \cellcolor{green!0}{40.7$\rightarrow$41.6} & \cellcolor{green!0}{54.3$\rightarrow$54.3} \\
        FT+Bi & \cellcolor{red!9}{62.1$\rightarrow$52.2} & \cellcolor{red!1}{39.7$\rightarrow$38.3} & \cellcolor{green!3}{75.0$\rightarrow$78.3} \\
        FT5 xl & \cellcolor{red!4}{51.1$\rightarrow$46.8} & \cellcolor{green!17}{36.6$\rightarrow$53.7} & \cellcolor{green!3}{39.3$\rightarrow$42.9} \\
                
         \bottomrule
    \end{tabular}
    \caption{Evaluation results of completeness permutation of Qwen2.5 7B (AO, CoT, FT and FT+Bi) and FlanT5 xl model (FT5 xl). Results are shown as \textbf{Original}$\rightarrow$\textbf{Remove First}.}
    \label{tab:rm_results}
\end{table}
Table~\ref{tab:encdec_em} shows the results from the non-finetuned encoder-decoder models. Generally, the Flan T5 family performs much better than non-finetuned decoder-only models for approximately similar parameter numbers. Flan T5 xl with 3B parameters already outperforms all decoder-only models under 8B with Answer Only or CoT, and achieves competitive performance compared to Qwen 2.5 14B. Interestingly, we can observe a very clear trend that the forward setting performs the best while the backward one performs the worst. The trend is clearer than in the Finetuned + Bi setting of the decoder-only models. 
When we tested other encoder-decoder models, the same trend was not observed, while the involved models continue to outperform equally-sized decoder-only models (see Table~\ref{tab:enc_dec_extra} in Appendix). We believe that this emerging ability of the Flan T5 models can be attributed to the selection of data used for their training \citep{10.5555/3618408.3619349}.

\subsection{Distance between Documents Matters}
\label{sec:distance}
We explore how the distance between gold documents affects the LMs' performance on MHQA using Qwen2.5 7B and FlanT5 xl. The results are shown in Figure~\ref{fig:distance}. Generally, the models' performance drops as the distance of the gold document increases. Notably, placing forward-ordered documents on the last positions of the input context brings significant performance improvement. This is because LMs generally favour documents close to border regions of the input prompt instead of the middle regions \citep{liu-etal-2024-lost}. Notably, finetuned models (Finetuned and Finetuned + Bi) remain more robust since their performance is less affected by both (i) the increase of distance between gold documents and (ii) the placement of forward-ordered documents at the last positions.

Our findings indicate that in a multi-iterative RAG setting ordering documents based on relevance rather than the order of their associated decomposed question (assuming that more than a single document is maintained for each decomposed question), can reduce the distance between relevant documents and effectively increase the end-to-end QA performance. As such, we emphasize the importance of incorporating ranking-based metrics to measure retrieval quality, which currently deviates from the standard practices in RAG-based MHQA, where the focus is primarily on recall (R@$n$) \cite{Trivedi2023, Gutierrez2024}.

\begin{figure*}[t]
    \centering    
    \begin{subfigure}[b]{0.32\textwidth} 
    \centering 
    \includegraphics[width=\linewidth]{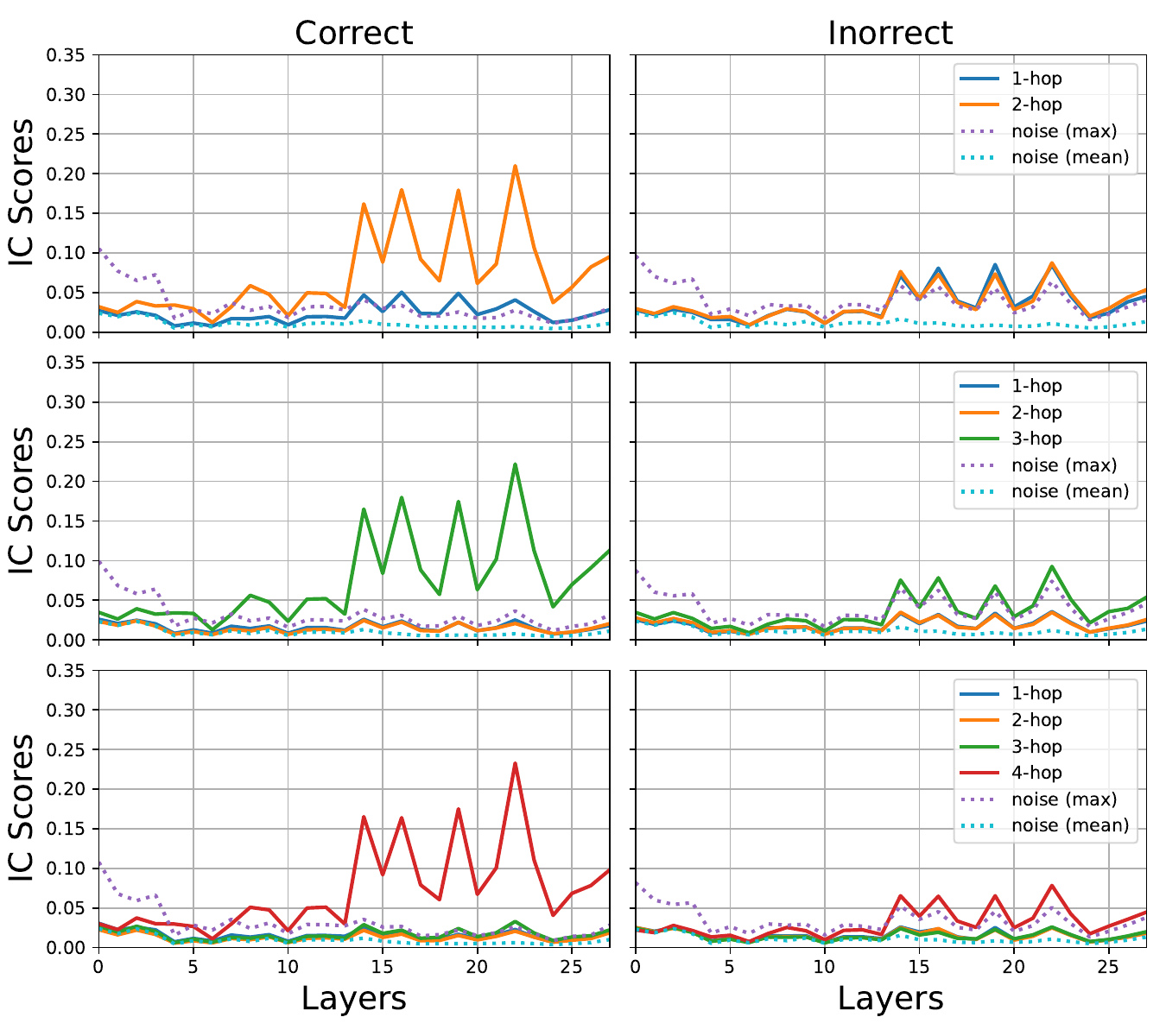}
    \caption{Answer Only}
    \end{subfigure} 
    \hfill
    \begin{subfigure}[b]{0.32\textwidth} 
    \centering 
    \includegraphics[width=\linewidth]{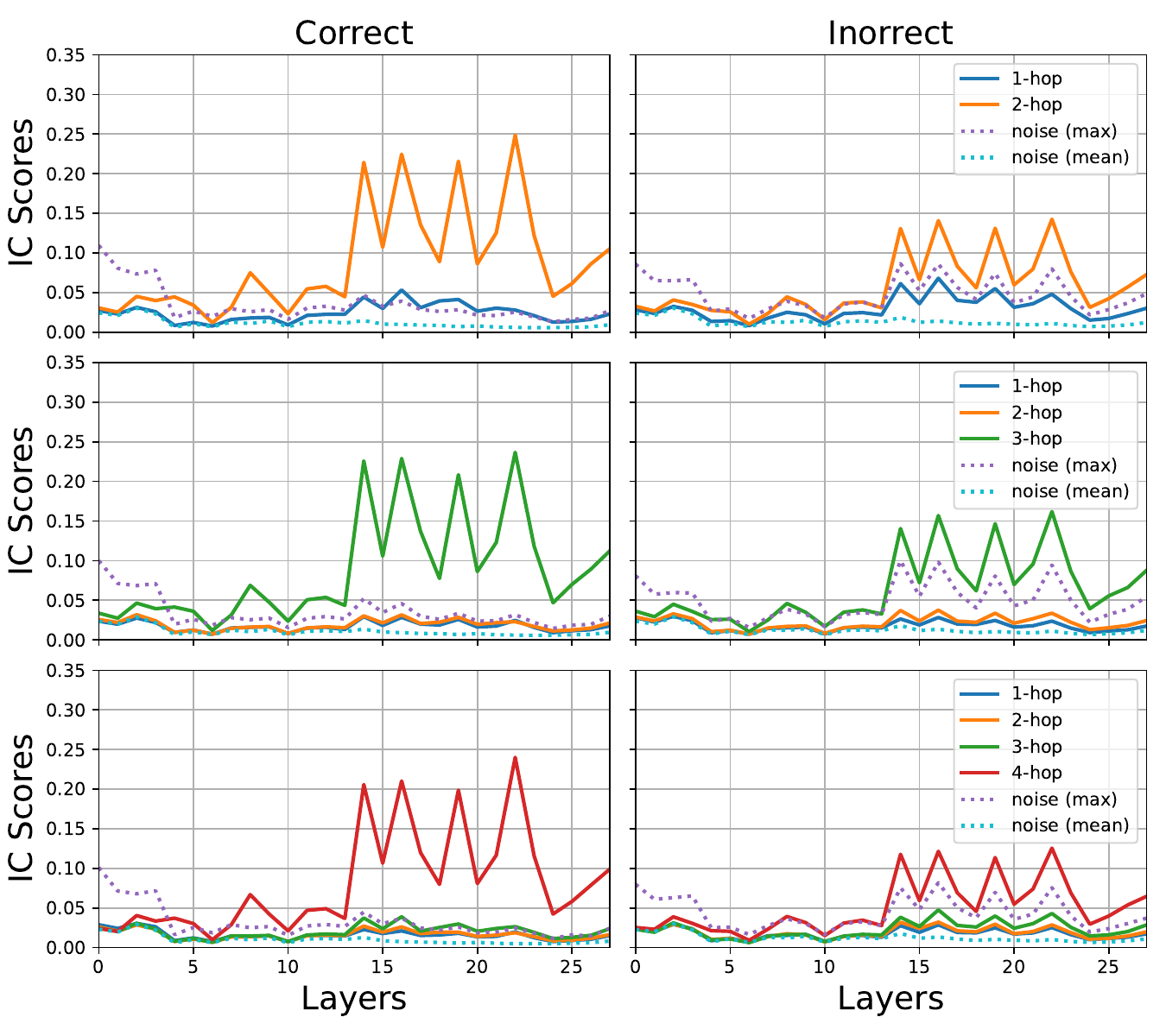}
    \caption{Finetuned}
    \end{subfigure} 
    \hfill
    \begin{subfigure}[b]{0.32\textwidth} 
    \centering 
    \includegraphics[width=\linewidth]{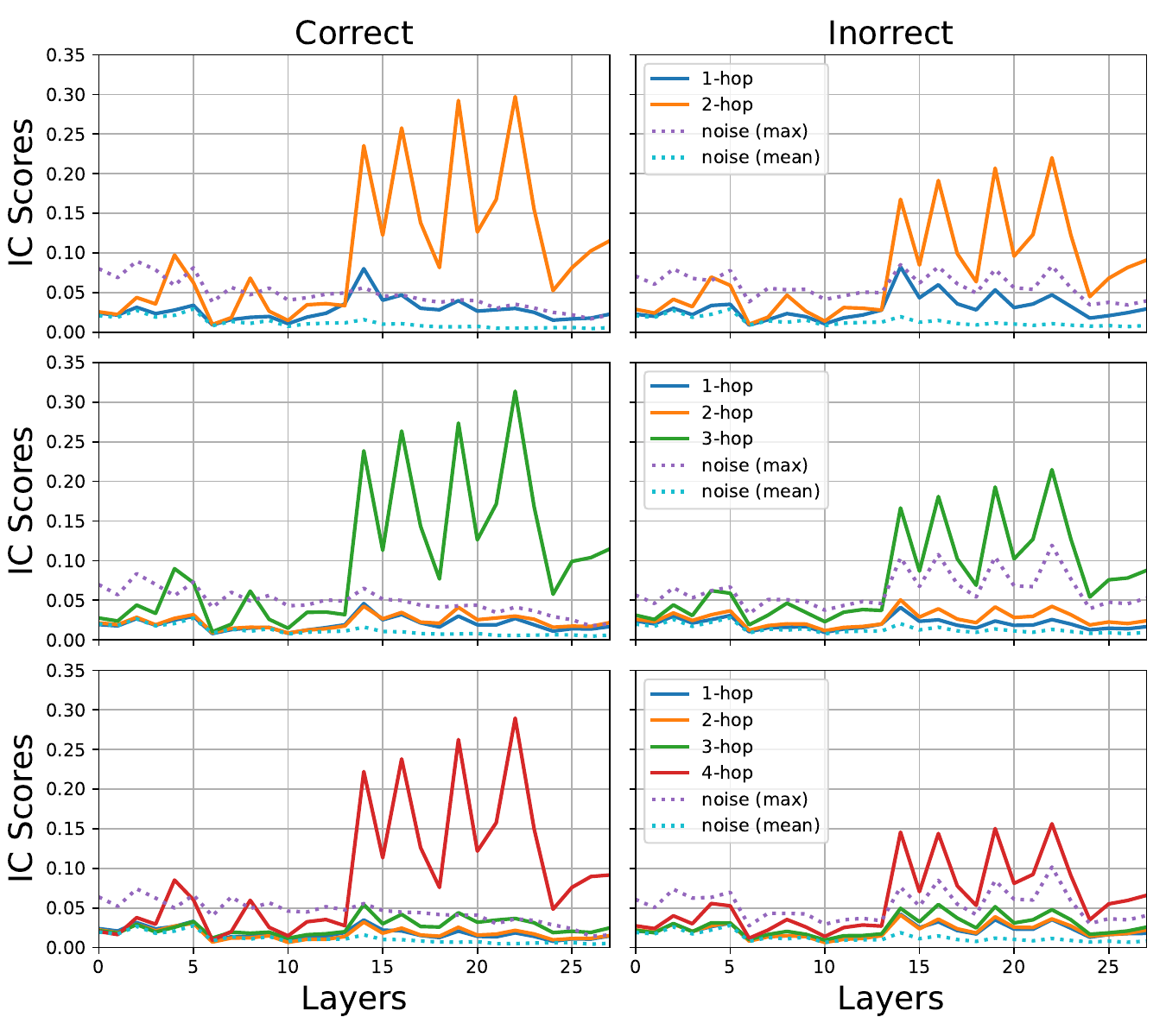}
    \caption{Finetuned + Bi}
    \end{subfigure} 

    \caption{IC distribution across different layers of Qwen2.5 7B with different setups for 2-hop, 3-hop and 4-hop questions, all in original order. LMs generally assign higher peak IC scores (in particular for the final hop) when answering the   multi-hop questions correctly.}
    \label{fig:attention_layer_general}
\end{figure*}

\begin{figure}[t]
    \centering
    \includegraphics[width=\linewidth]{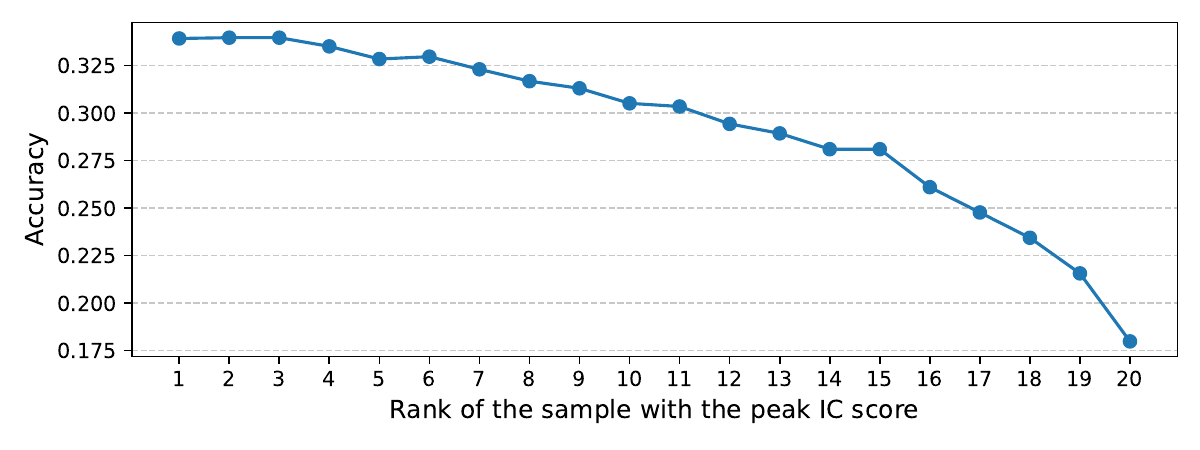}
    \caption{Qwen 2.5 7B performance (Answer Only) peak IC ranking from higher peak IC to lower peak IC. A clear trend can be observed that higher peak IC among the 20 random shuffles brings higher accuracy.}
    \label{fig:acc_ic}
\end{figure}

In Appendix~\ref{appendix_sec:2wiki_experiments}, we include additional experiments on the 2WikiMultihopQA \texttt{Compositional} subset. The findings are in line with the major observations above in terms of the distance and order of the gold documents.

\subsection{Do LMs guess the answer?}
\label{sec:rm_first}
To identify if LMs can answer questions even when they do not have enough information, we manually remove the first hop document from the context. Since LMs may have the relevant parametric knowledge already, we further design a simple atomic by asking them the first hop question, without any other context. We determine whether they have the required parametric knowledge by evaluating their answer. Table~\ref{tab:rm_results} shows the relevant results. 

\paragraph{Without first-hop information in parametric knowledge}
For 2-hop questions, since the first hop document is more important, generally all models across all settings drop in accuracy when the information is not stored in their parameters. For 3-hop and 4-hop questions, we see a similar performance drop on non-finetuned models, but the drop is less significant. This means for complex questions, LMs still lack the ability to know what they don't know, and they are expected to refuse to answer the question since the reasoning path of existing evidence is not complete. For finetuned models, the issue is more severe, where the accuracy even increases (4-Hop, FT) as key information is not provided.

\paragraph{With first-hop information in parametric knowledge}
Similarly, for 2-hop questions, accuracy drops in most settings, but it is not as significant as when the removed information is known by the model. While for more complex questions, performance generally increased, especially for the CoT and Flan T5 models. Even when we ask models to only use the given context, they still default to parametric knowledge for MHQA. This also suggests that complex MHQA is still a big challenge to LMs, where failure to locate all required information in the context may lead to hallucinations. This also indicates that retrieval may sometimes harm models' performance, even if the external knowledge does not conflict with parametric knowledge.

\section{Are LMs Aware of Context Permutations?}
To investigate how LMs make use of the given context to answer multi-hop questions, 
we compute the information contribution (IC) score with Equation~\eqref{equ:ic} on the MuSiQue development set.
Our hypothesis is that by investigating how models \textit{pay attention} to particular documents in the context, we can better explain their behaviour in answering multi-hop questions. In this part, we only consider three settings: \textbf{Answer Only}, \textbf{Finetuned} and \textbf{Finetuned + Bi}. Figure~\ref{fig:attention_layer_general} shows the IC distribution of the different setups across different layers of the Qwen 7B model. %

\subsection{LMs Assign Higher Peak Attention when Correct}
\label{sec:peak_behave}
For the samples that are answered correctly, the model consistently assigns the largest attention score to the last hop document. This is intuitive since the answer is included in the last hop document. In the case of samples that are answered incorrectly, we observe a smaller gap between the last hop and previous hop documents. Notably, while correct samples always assigned higher attention scores to the gold documents, the IC score of the Answer Only setup with direct prompting is generally lower than the two settings based on finetuning. This is even clearer for the incorrect samples, where, for the non-finetuned variants, the highest IC score of the noise documents (labelled as noise (max)) is almost the same as that of the gold documents.  Interestingly, the ``retrieval'' layers to which most of the fluctuations in the attention weight are attributed remain consistent throughout fine-tuning, even with bi-directional attention.

We find that all models assign much higher attention to the last hop when generating the correct answer, but are less confident and assign more evenly distributed attention scores when generating incorrect answers. More generally, all models assign higher peak attention (i.e., largest attention to one document in the context) when they predict the answer correctly. To better understand the relation between model predictions and the largest IC score, we randomly shuffle the document order in the context 20 times per question, then perform inference with Qwen2.5 7B in the Answer Only setup. We observe that the median peak IC scores are 2.22 and 1.72 for correct and incorrect samples (cf. Figure~\ref{fig:boxplot_ics_7b_exbox}). %
IC scores are in general higher for correct than incorrect samples. 

In addition, we compute the prediction accuracy with different context shuffles based on the ranked peak IC scores (Figure~\ref{fig:acc_ic}). Our results show that a higher peak IC score among the 20 random shuffles provides higher MHQA performance, showcasing that the peak IC score is a key signal from the LM to identify the optimal context order. %
This finding underscores the importance of context order for the MHQA task, where the best context order almost doubles performance over the worst order.

\subsection{LMs Generally Favour the Last Document}
Amongst all noise documents, we capture the most favoured position where the IC score is the highest. We find that the last document receives the most attention in most cases. In Figure~\ref{fig:attention_layer_general}, the purple dashed line shows the attention assigned to the noise that confuses the model most. Interestingly, we find this is mostly the last document, which seems to have a high probability of being captured in the lower layer of the non-finetuned model. In Figure~\ref{fig:attention_layer_general}, there is always a local peak of noise document at early layers.

\begin{figure}[t]
    \centering
    \includegraphics[width=\linewidth]{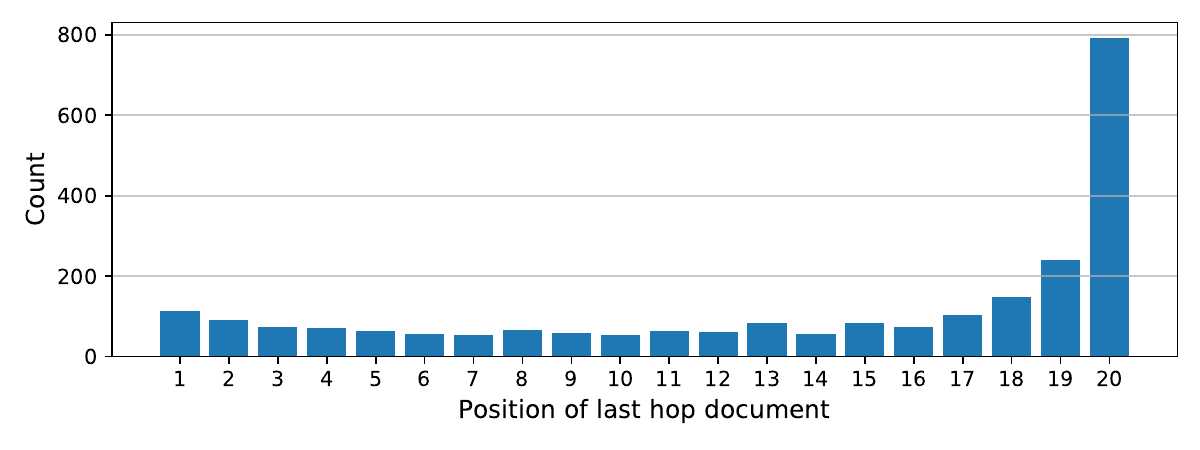}
    \caption{Position of last hop document when the order achieves highest peak IC score in Qwen 7B model (Answer Only). LMs generally assign highest peak IC when the last hop document is close to the end.
    }
    \label{fig:last_hop_7b}
\end{figure}

In addition, in all the best samples (with the largest IC score among 20 randomly shuffled contexts) from Section~\ref{sec:peak_behave}, we also observe a particular preference for the last document (cf. Figure~\ref{fig:last_hop_7b}). A similar phenomenon is observed in other causal decoder-only models, even for models like Qwen2.5 0.5B which performs worst on forward order, the trend exists whilst less significant. 
\section{Discussion}

\subsection{Optimizing LMs' Use of Context}
State-of-the-art RAG methods \citep{Trivedi2023, Gutierrez2024} for solving multi-hop question generally split the $n$-hop complex question $q$ into several decomposed questions $q_i$ and retrieve top-$k$ documents $D_{i}=\{d_{i,1}, d_{i,2}, ..., d_{i,k}\}$ accordingly. Simply concatenating the documents with the order of decomposed questions  
ensures that the document order is forward, but the distance between relevant documents is large. According to our observations in Section~\ref{sec:order} and~\ref{sec:distance}, non-finetuned causal language models are not sensitive to the relevant documents' order, but the distance between relevant documents matters. Thus if the selected reader is an off-the-shelf causal language model, gathering additional documents, \textit{ensuring that their in-between distance is minimum}, is essential. On the contrary, if finetuning is an option, then keeping the forward order is more important for getting the best performance. For both situations, it is strongly recommended to place documents with higher relevance close to the end of the context.

\subsection{Use Reader with Bidirectional Attention}
Causal decoder-only LMs are widely used in the RAG framework as readers. In our experiments, we show that these models are not the ideal choice as MHQA solvers, as they are limited by their causal mask. Our experiments show that by altering the causal mask with a prefix mask, and simply using LoRA finetuning to obtain a non-causal decoder-only model, we can outperform the original causal setup while being more robust against the order of gold documents. Notably, the FlanT5 family shows marvellous off-the-shelf MHQA ability and can serve as a competitive reader alternative to causal decoder-only models within the RAG framework. 

\subsection{Attribution Is Important}
From Section~\ref{sec:rm_first}, we observe that removing the first hop document causes a performance drop to all language models, but still good accuracy is maintained. In addition, we find that the attention weight assigned to the last hop document does not change when removing the first hop document (Figure~\ref{fig:attention_layer_rm}). These findings further underscore the importance of attribution in RAG, to ensure that the predicted answer is supported by evidence from the context.

\section{Conclusion}
In this study, we explore the MHQA capabilities of LMs with different architectures. We find that non-finetuned, causal decoder-only LMs are invariant to the order of gold documents but are affected by their in-between distance in the input context. Incorporating bi-directional attention enhances performance---both in the case of encoder-decoder models, which outperform their decoder-only counterparts and when applied to decoder-only models. Additionally, finetuning instils in the models the bias of forward order and makes them more robust against the distance between gold documents.
Our analysis of attention distribution indicates that increased peak attention in the context aligns with accurate predictions. These insights advance our understanding of LMs in MHQA and suggest avenues for future improvements.

\section*{Acknowledgments}
This work is supported by UKRI (grant number EP/S022481/1), The University of Edinburgh and  Huawei’s Dean’s Funding (C-00006589).

\section*{Limitations}
In this work, with the limitation of computing resources, we consider MuSiQue dataset with its original setup, with at most 20 documents per question. To the best of our knowledge, MuSiQue is one of the most challenging datasets for MHQA, and, therefore, ideal for the experiment we want to conduct in this study. In Appendix \ref{appendix_sec:2wiki_experiments}, we include additional results on relevant subsets of 2WikiMultihopQA, demonstrating that our original findings on MuSiQue hold.

Most of our prompts are around or below 4k tokens, which is relatively a short setting with respect to the current long-context language models. We believe that this does not diminish the contribution of this work. For example, we have already noticed a significant effect of distance between gold documents with only 5 noise documents in between, indicating that when considering longer contexts, the context order is even more important for complex reasoning tasks, such as MHQA, which is a crucial issue that needs to be considered in future works.

\bibliography{anthology,custom}

\appendix
\section{Experiment Environment}
All the experiments mentioned in this paper are undertaken on NVIDIA A100 80GB GPUs. To make sure that our experiments are replicable, we utilize greedy decoding for all the inference experiments and set the seed to 42 for all finetuning experiments. We use Huggingface Transformers library \citep{wolf-etal-2020-transformers} to accomplish all the experiments in this study, and utilize PEFT \citep{peft} for LoRA finetuning.

\section{Model Specification}
Table~\ref{tab:llm_spec} shows the specifications of LMs investigated in this study. 
\label{sec:appendix}

    \begin{table}[]
        \centering
        \small
        \begin{tabular}{l|cccc}
        \toprule
            \textbf{Model} & \textbf{Params} & \textbf{Layers} & \textbf{Dim} & \textbf{Heads} \\ \midrule
            Qwen2.5 0.5B & 0.49 & 24 & 896 & 14/2 \\
            Qwen2.5 1.5B & 1.5 & 28 & 1,536 & 12/2 \\
            Qwen2.5 3B & 3.1 & 36 & 2,048 & 16/2 \\
            Qwen2.5 7B & 7.6 & 28 & 3,584 & 28/4 \\
            Qwen2.5 14B & 14.7 & 48 & 5,120 & 40/8 \\ \midrule
            Llama3.2 1B & 1.23 & 16 & 2,048 & 32/8 \\
            Llama3.2 3B & 3.21 & 28 & 3,072 & 24/8\\
            Llama3.1 8B & 8.03 & 32 & 4,096 & 32/8 \\ \midrule
            Flan T5 small & 0.08 & 8+8 & 512 & 6 \\
            Flan T5 base & 0.25 & 12+12 & 768 & 12 \\
            Flan T5 large & 0.8 & 24+24 & 1,024 & 16 \\
            Flan T5 xl & 3 & 24+24 & 2,048 & 32 \\
            Flan T5 xxl & 11 & 24+24 & 4,096 & 64 \\
            
        \bottomrule
        \end{tabular}
        \caption{Specification of LMs experimented in this paper.}
        \label{tab:llm_spec}
\end{table}

\section{Prompt Details}
The prompt used in this work is adapted from \citet{liu-etal-2024-lost} as:
\begin{tcolorbox}
    Answer the question using only the provided search results (some of which might be irrelevant).
    \newline
    \newline
    \textcolor{red}{$\langle$Documents$\rangle$}
    \newline
    \newline
    Question: \textcolor{red}{$\langle$Question$\rangle$}
\end{tcolorbox}

Specially, for \textbf{Answer Only} setup, we explicitly add \texttt{\textbackslash boxed\{} to the end of the prompt to force model following the answer only instruction. 

\section{Detailed statistics of peak IC score}
\begin{figure}[t]
    \centering
    \includegraphics[width=\linewidth]{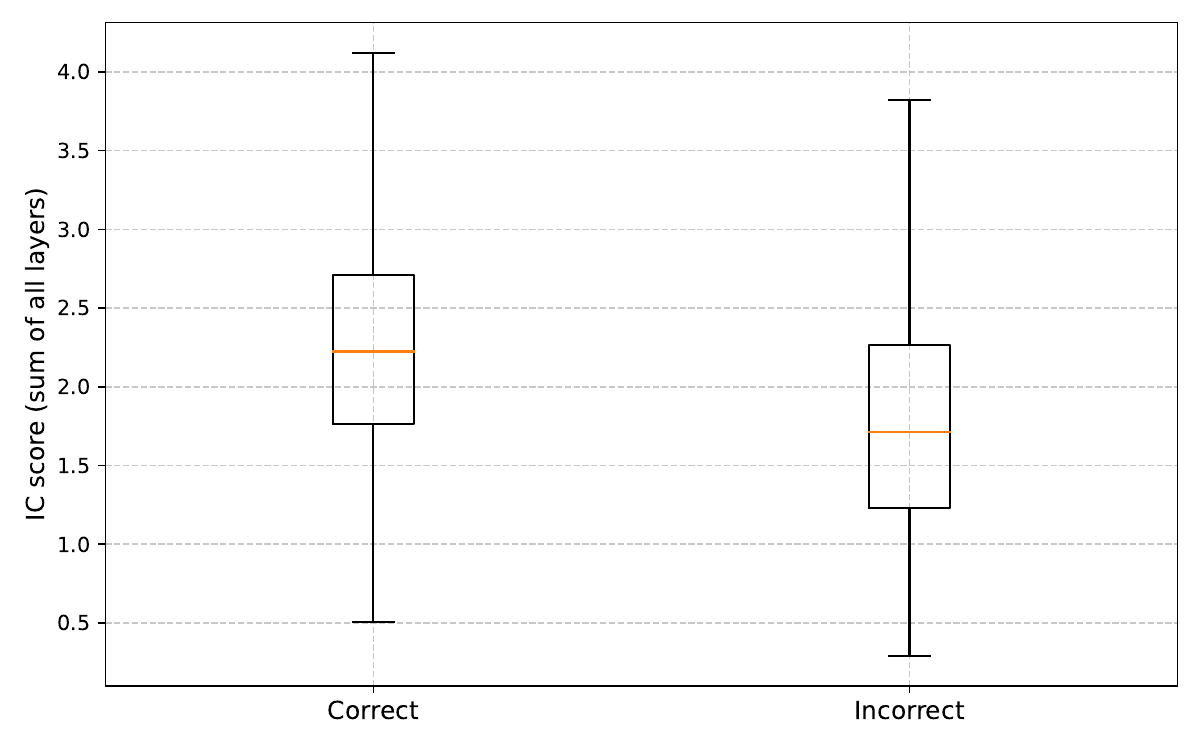}
    \caption{Statistics of peak IC score of correct and incorrect samples.}
    \label{fig:boxplot_ics_7b_exbox}
\end{figure}

As discussed in Section~\ref{sec:peak_behave}, we observe that LMs assign higher peak IC scores when they get the correct answer. Figure~\ref{fig:boxplot_ics_7b_exbox} shows the box plot of peak IC score when correct and incorrect answers are predicted, with 20 random shuffles for each question. 
\section{Extra analysis of IC plots}
\subsection{Model behave differently with Backward and Forward}
\begin{figure*}[t]
    \centering    
    \begin{subfigure}[b]{0.16\textwidth} 
    \centering 
    \includegraphics[width=\linewidth]{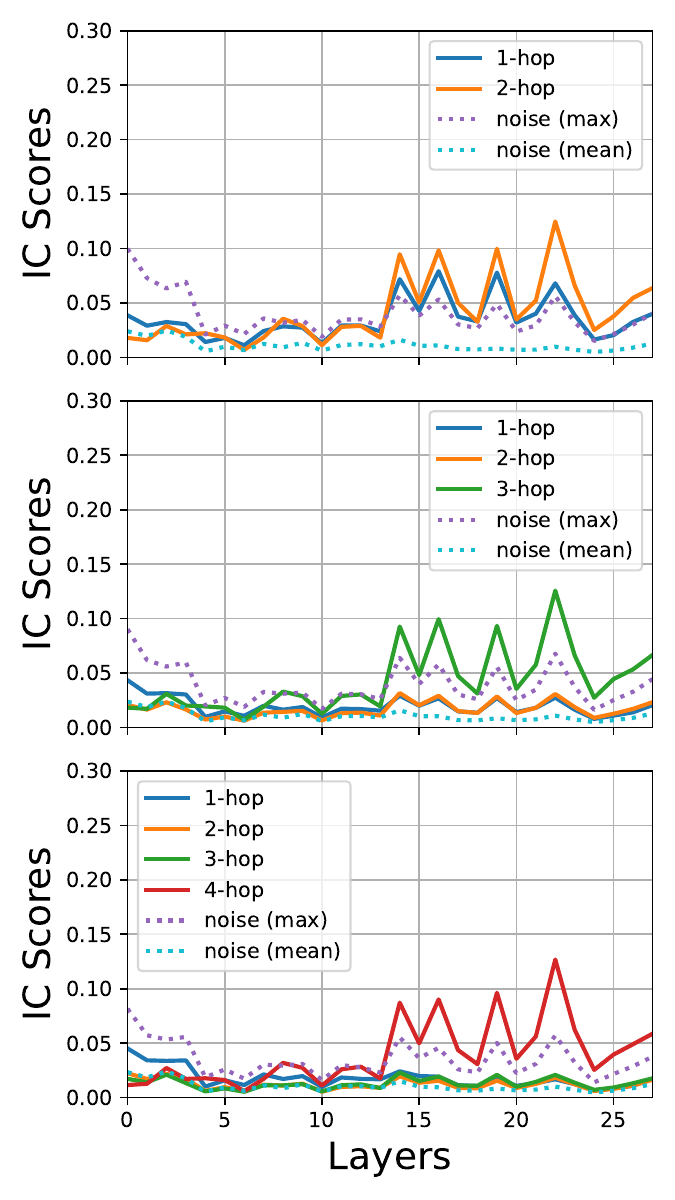}
    \caption{AO, Backward}
    \label{fig:attention_layer_order:a}
    \end{subfigure} 
    \hfill
    \begin{subfigure}[b]{0.16\textwidth} 
    \centering 
    \includegraphics[width=\linewidth]{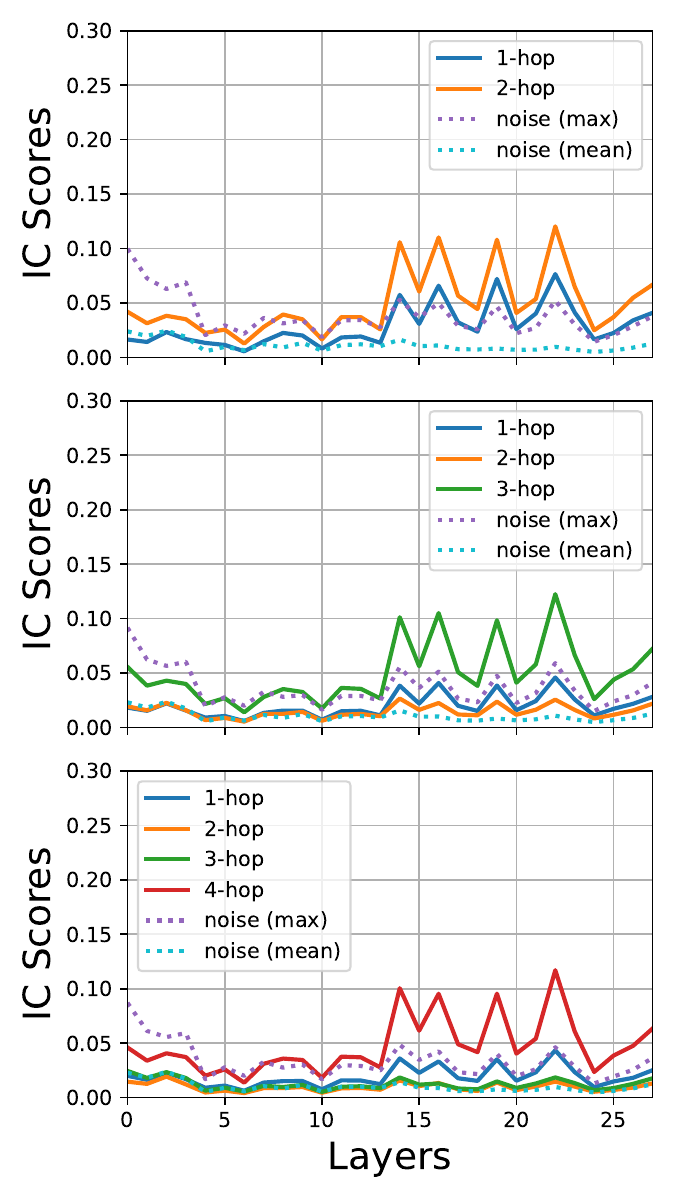}
    \caption{AO, Forward}
    \label{fig:attention_layer_order:b}
    \end{subfigure} 
    \hfill
    \begin{subfigure}[b]{0.16\textwidth} 
    \centering 
    \includegraphics[width=\linewidth]{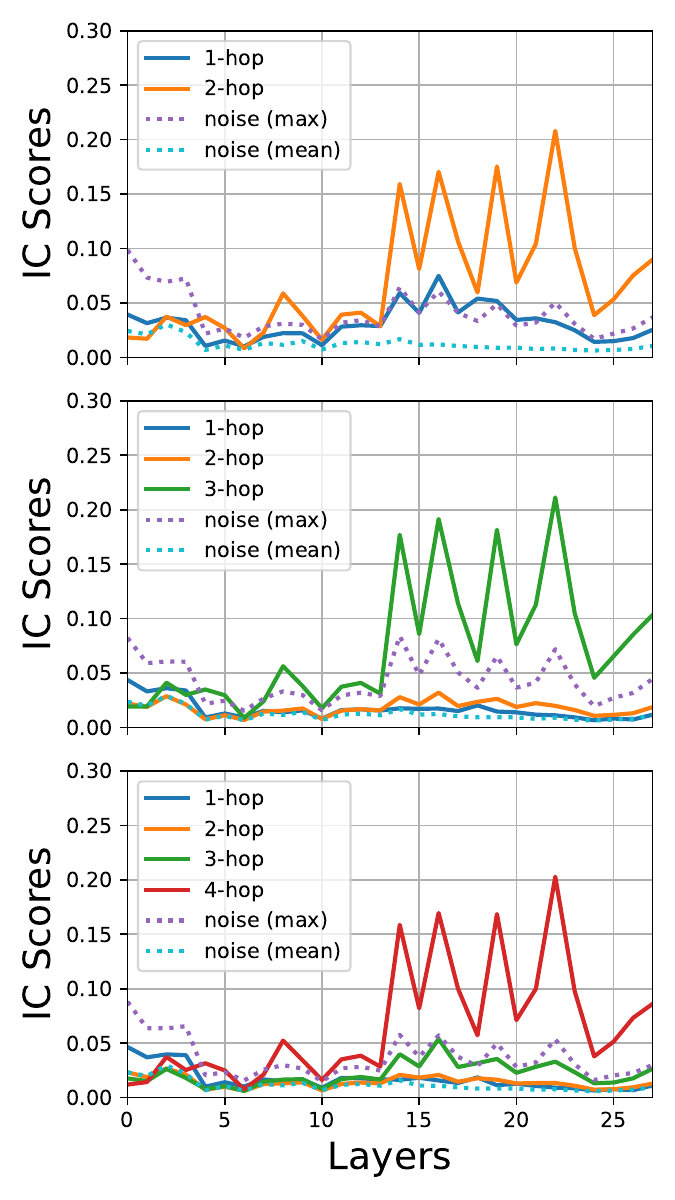}
    \caption{FT, Backward}
    \end{subfigure} 
    \hfill
    \begin{subfigure}[b]{0.16\textwidth} 
    \centering 
    \includegraphics[width=\linewidth]{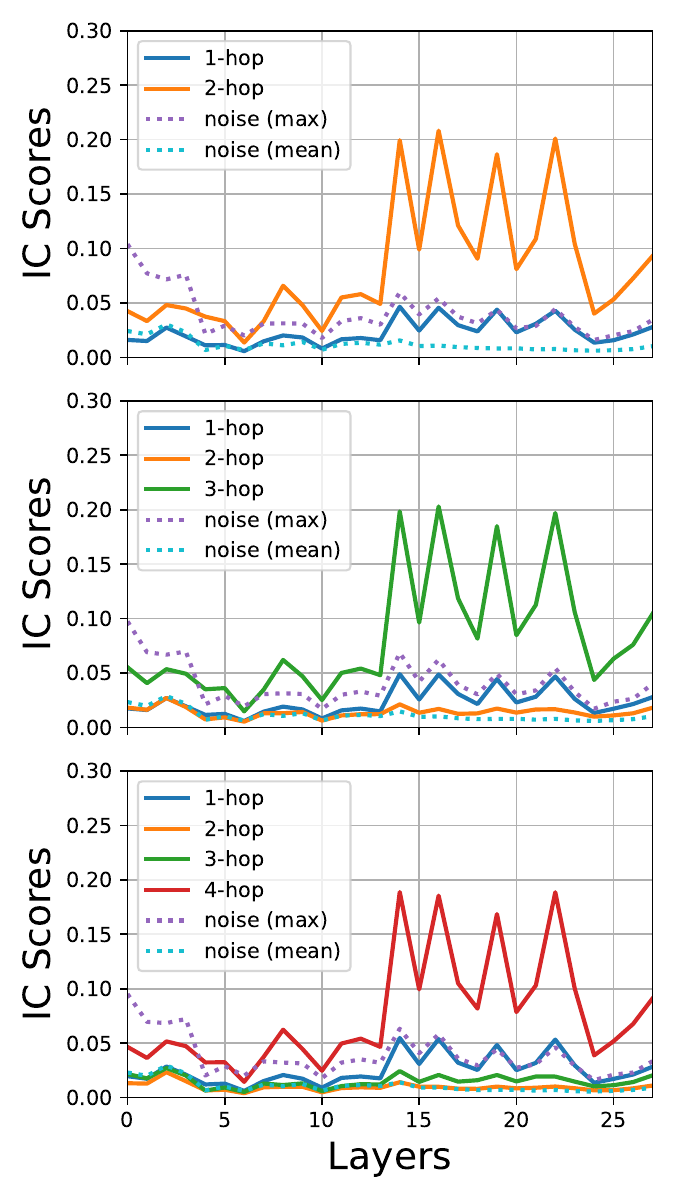}
    \caption{FT, Forward}
    \end{subfigure} 
    \hfill
    \begin{subfigure}[b]{0.16\textwidth} 
    \centering 
    \includegraphics[width=\linewidth]{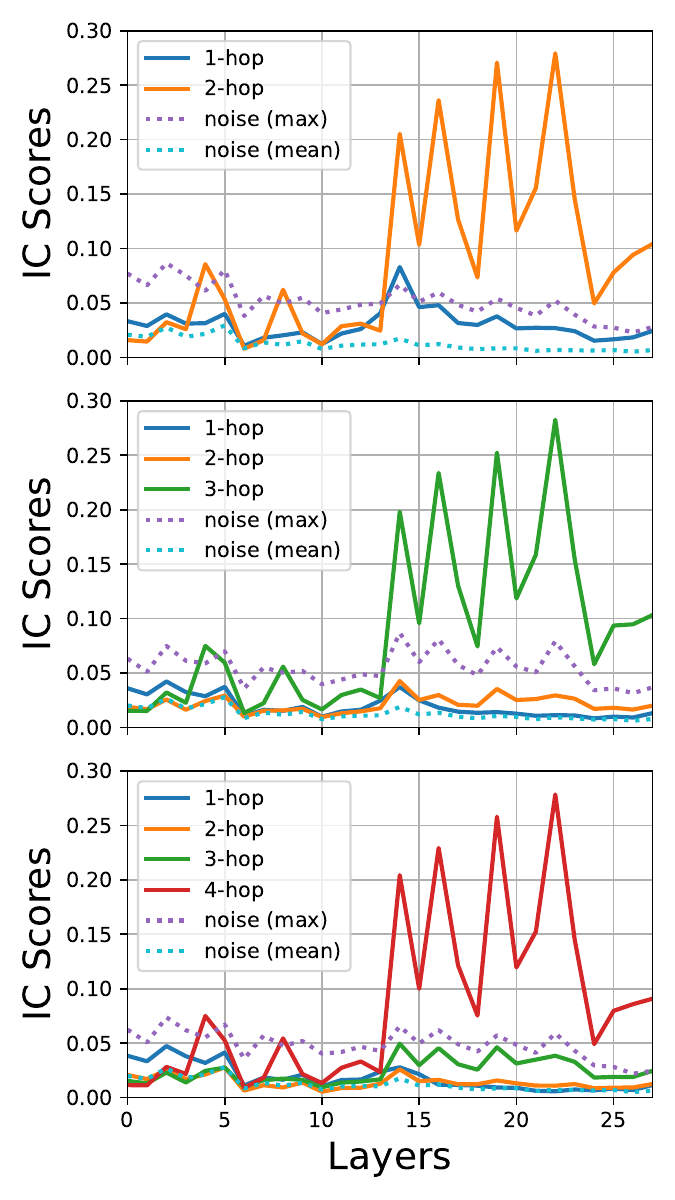}
    \caption{Bi, Backward}
    \end{subfigure} 
    \hfill
    \begin{subfigure}[b]{0.16\textwidth} 
    \centering 
    \includegraphics[width=\linewidth]{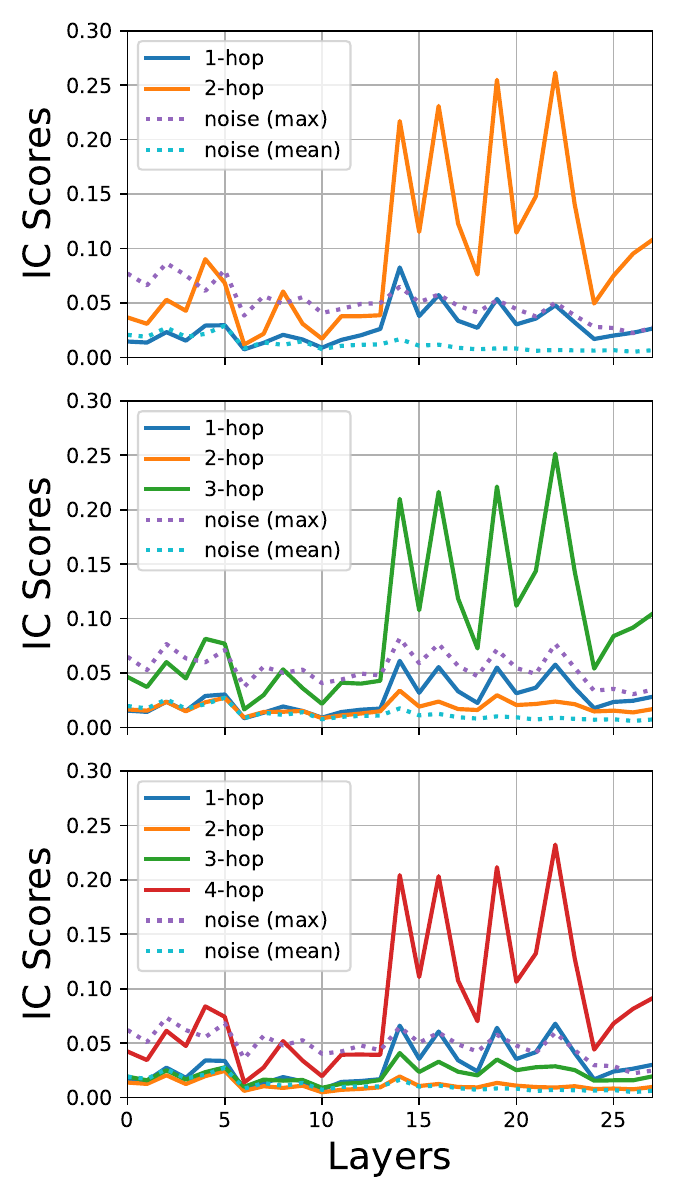}
    \caption{Bi, Forward}
    \end{subfigure} 

    \caption{IC distribution across different layers of Qwen2.5 7B with different order of gold document.}
    \label{fig:attention_layer_order}
\end{figure*}

In this section, we illustrate other findings observed in the IC plots.
\label{sec:behave_order}
From Figure~\ref{fig:attention_layer_order}, we can find that in the Forward setup, the model tends to focus more on the first hop document. For instance, in the IC distribution on 4-hop questions of Figure~\ref{fig:attention_layer_order:a} and \ref{fig:attention_layer_order:b}, the blue plot of attention focusing on a 1-hop document in Forward setup is the highest except the last hop, while in Backward setup the differences between these hops are minor. This behaviour also generalizes to finetuned models. Moreover, this behaviour is mainly observed in 3-hop and 4-hop questions, where the distance between the first-hop document and last-hop document is longer. Though last hop document is able to encoding part of the information of first hop document, longer distance will reduce this encoding and requires model to focus more on it to prevent ``forgetting''.

\subsection{How does distance change the LM's mind?}
\begin{figure*}[t]
    \centering    
    \begin{subfigure}[b]{0.16\textwidth} 
    \centering 
    \includegraphics[width=\linewidth]{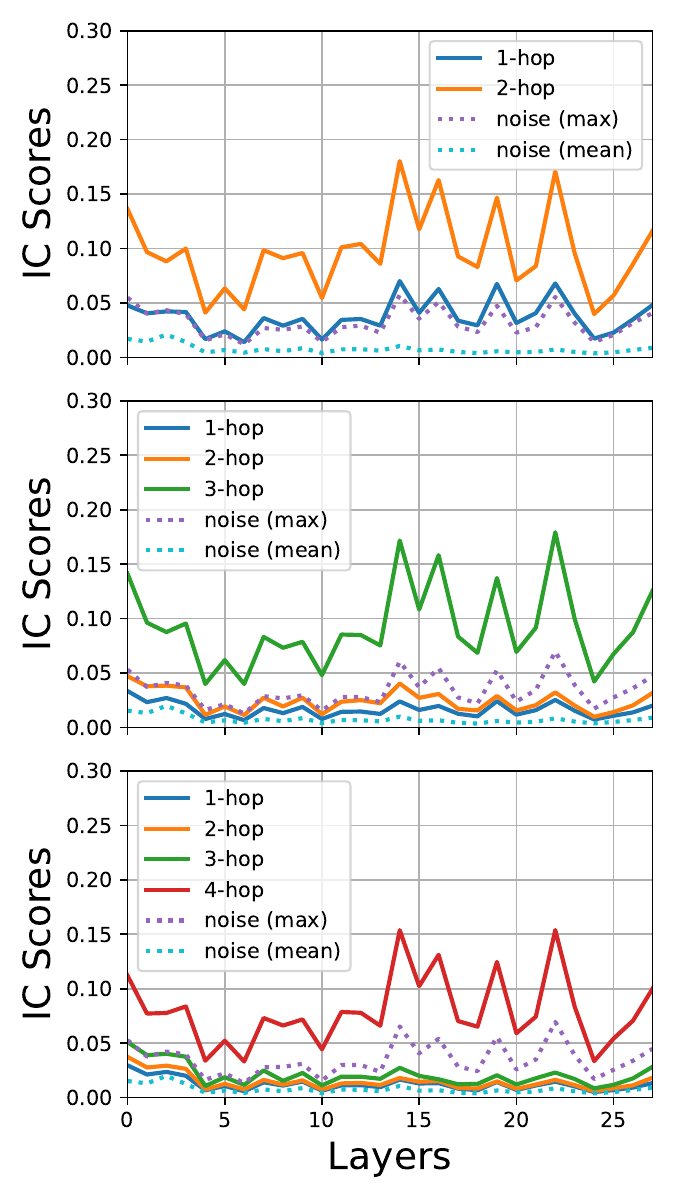}
    \caption{AO, $i$=0}
    \end{subfigure} 
    \hfill
    \begin{subfigure}[b]{0.16\textwidth} 
    \centering 
    \includegraphics[width=\linewidth]{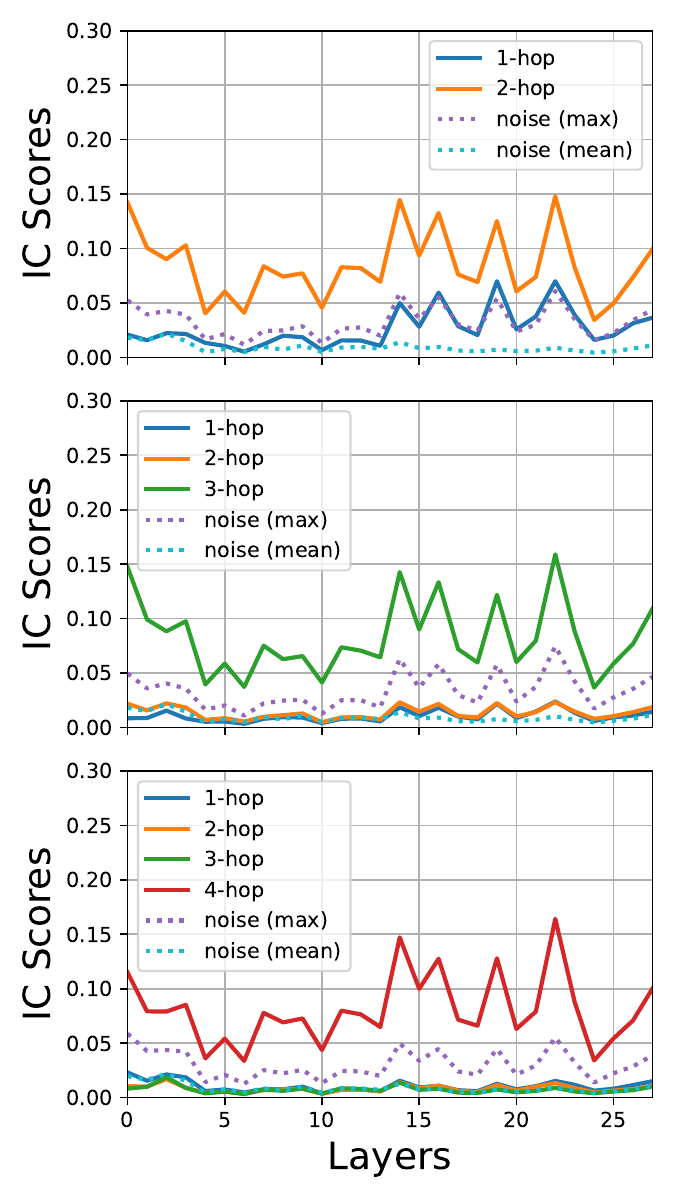}
    \caption{AO, $i$=5}
    \end{subfigure} 
    \hfill
    \begin{subfigure}[b]{0.16\textwidth} 
    \centering 
    \includegraphics[width=\linewidth]{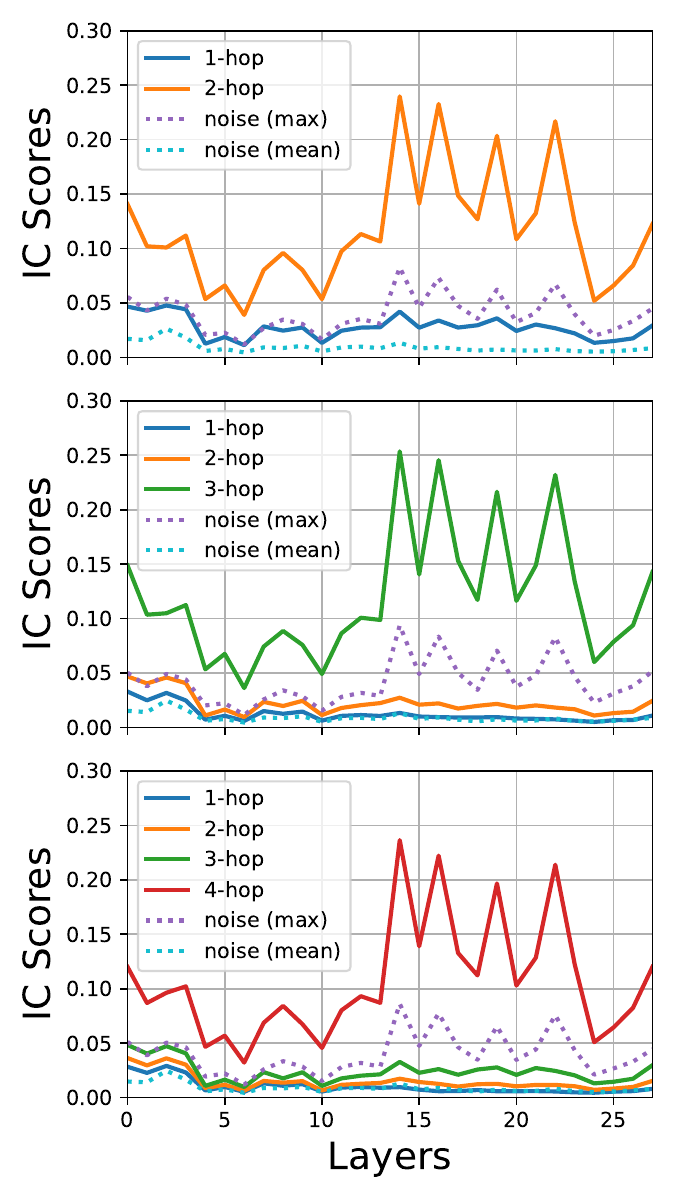}
    \caption{FT, $i$=0}
    \end{subfigure} 
    \hfill
    \begin{subfigure}[b]{0.16\textwidth} 
    \centering 
    \includegraphics[width=\linewidth]{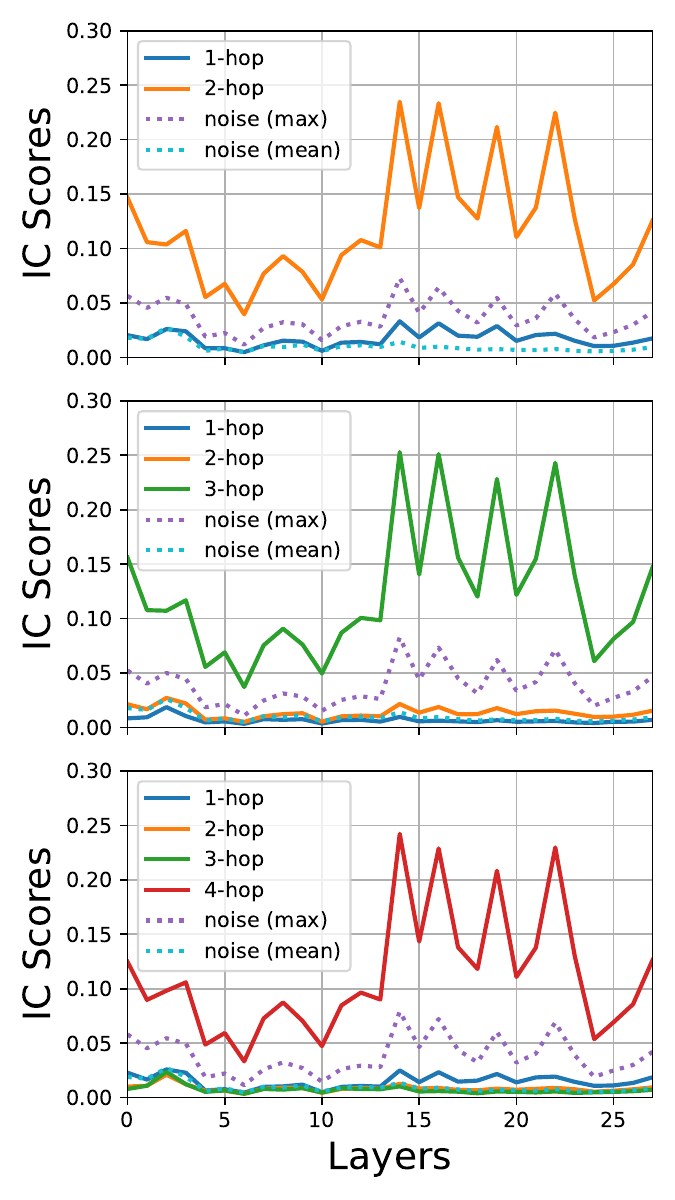}
    \caption{FT, $i$=5}
    \end{subfigure} 
    \hfill
    \begin{subfigure}[b]{0.16\textwidth} 
    \centering 
    \includegraphics[width=\linewidth]{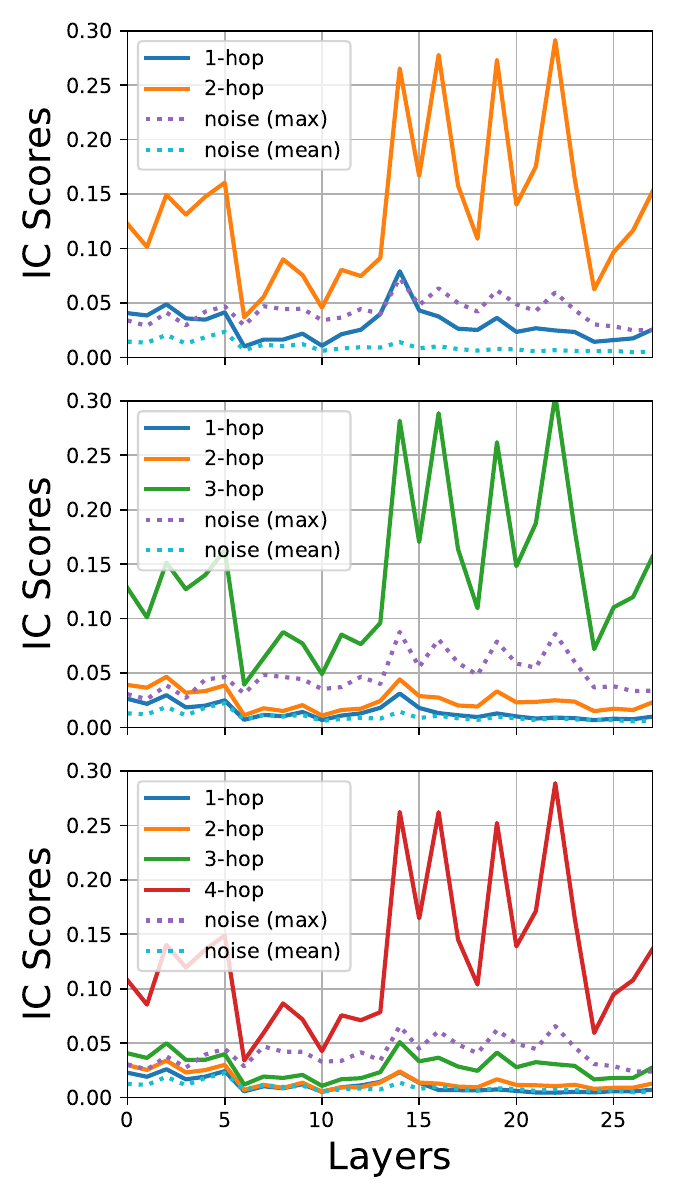}
    \caption{FT+Bi, $i$=0}
    \end{subfigure} 
    \hfill
    \begin{subfigure}[b]{0.16\textwidth} 
    \centering 
    \includegraphics[width=\linewidth]{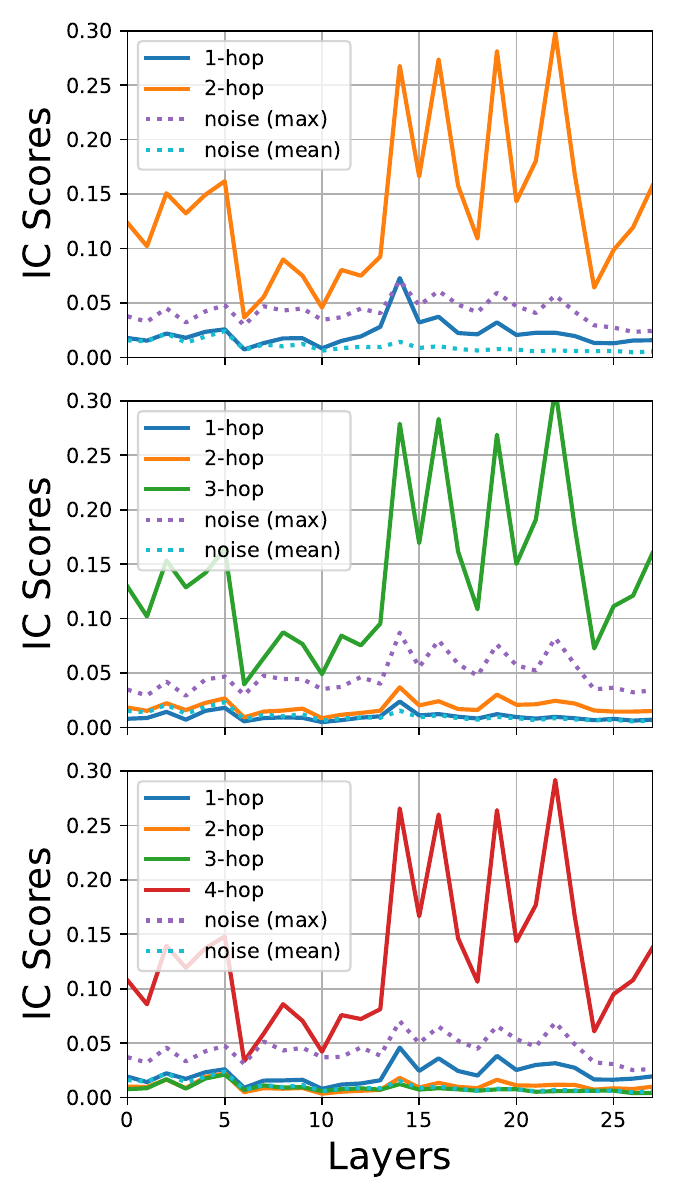}
    \caption{FT+Bi, $i$=5}
    \end{subfigure} 

    \caption{IC distribution across different layers of Qwen2.5 7B with different distances $i$ between gold documents.}
    \label{fig:attention_layer_distance}
\end{figure*}

Figure~\ref{fig:attention_layer_distance} shows the IC distribution across layers, considering different setups and distances (i.e., $i$ in Forward\_$i$ settings) between gold documents. 
Specifically, when $i=5$, the model behaves similarly as discussed in Section~\ref{sec:behave_order}: the first hop document gets the most attention except the last hop, but when $i=0$, the attention intensity starts to follow the order of the reasoning hops, which decreases from the last hop to the first hop. This finding supports our hypothesis in Section~\ref{sec:behave_order}, suggesting that a longer distance compels the model to allocate more attention to the first-hop document to ensure essential information is retained when gold documents are forward-placed.

\subsection{How does removing the first hop change the LM's mind?}
\begin{figure*}[t]
    \centering    
    \begin{subfigure}[b]{0.16\textwidth} 
    \centering 
    \includegraphics[width=\linewidth]{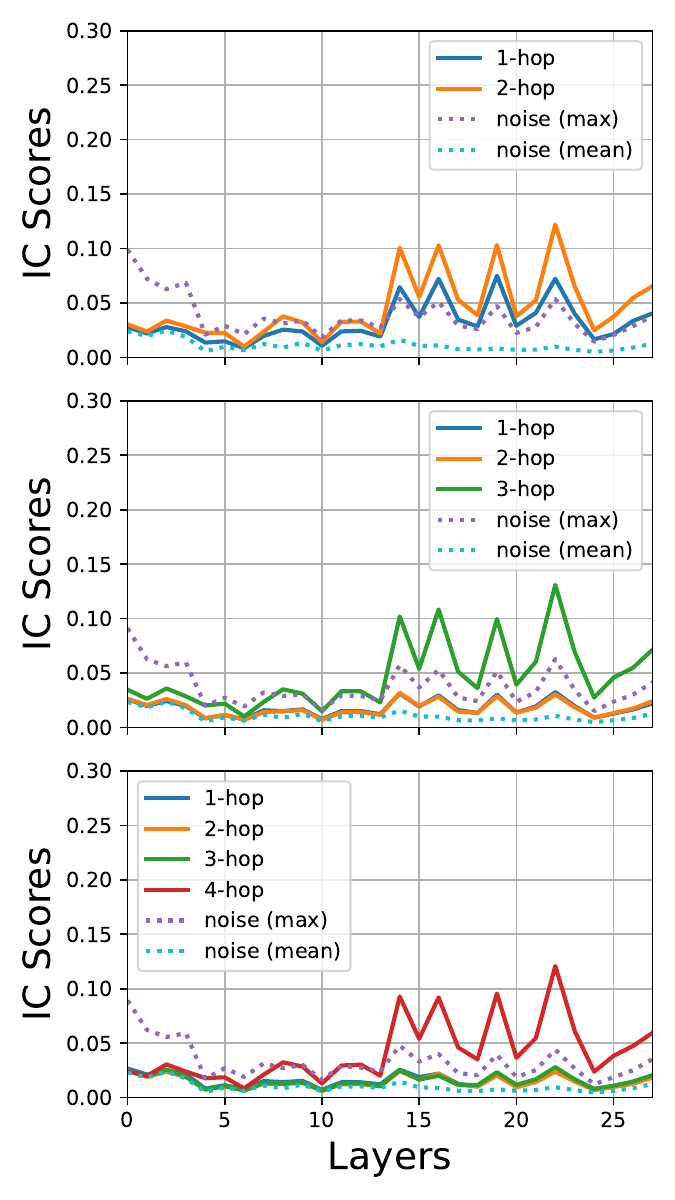}
    \caption{AO, w/ 1 hop}
    \end{subfigure} 
    \hfill
    \begin{subfigure}[b]{0.16\textwidth} 
    \centering 
    \includegraphics[width=\linewidth]{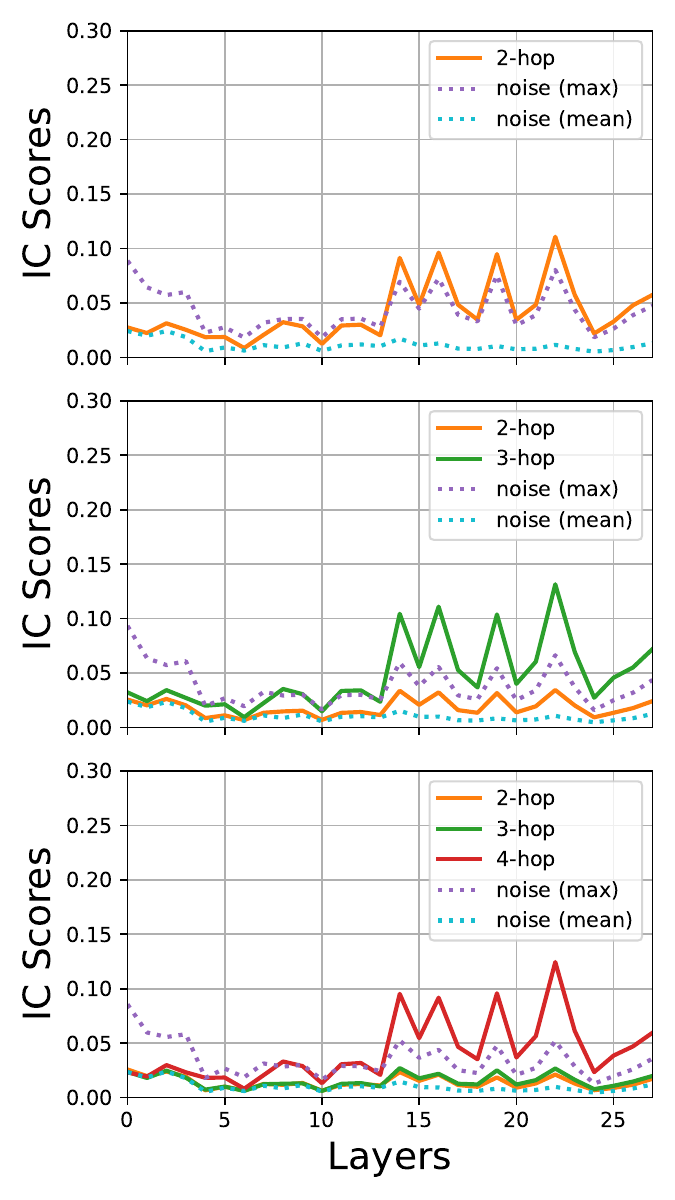}
    \caption{AO, w/o 1 hop}
    \end{subfigure} 
    \hfill
    \begin{subfigure}[b]{0.16\textwidth} 
    \centering 
    \includegraphics[width=\linewidth]{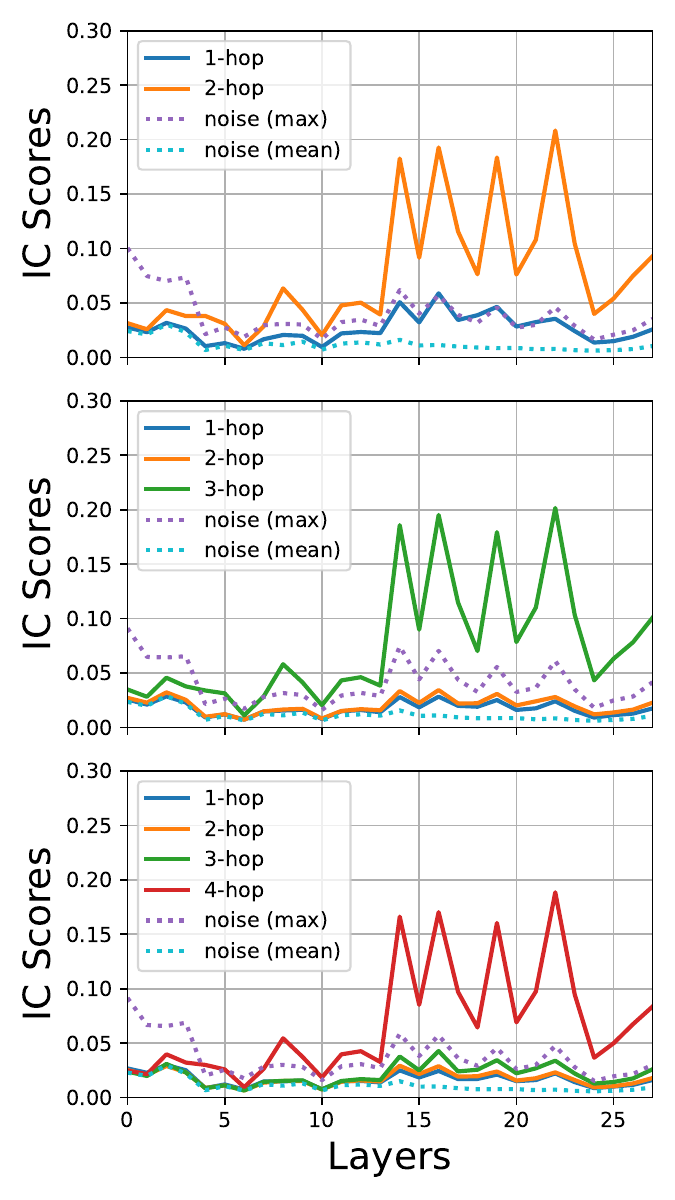}
    \caption{FT, w/ 1 hop}
    \end{subfigure} 
    \hfill
    \begin{subfigure}[b]{0.16\textwidth} 
    \centering 
    \includegraphics[width=\linewidth]{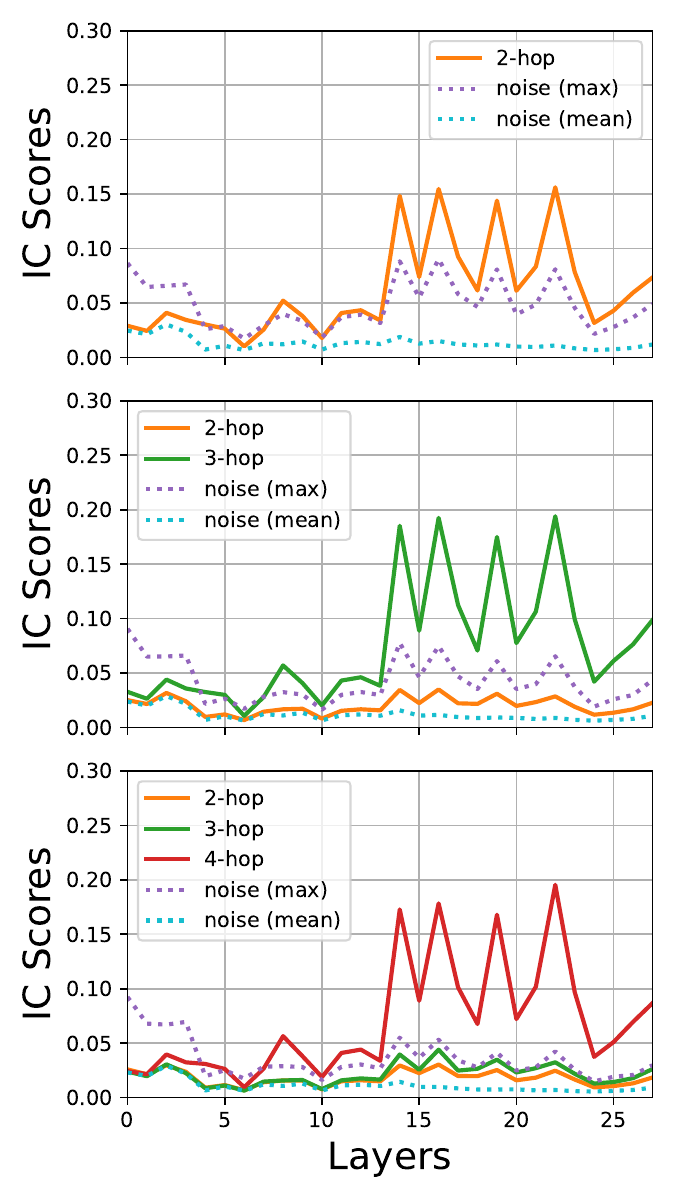}
    \caption{FT, w/o 1 hop}
    \end{subfigure} 
    \hfill
    \begin{subfigure}[b]{0.16\textwidth} 
    \centering 
    \includegraphics[width=\linewidth]{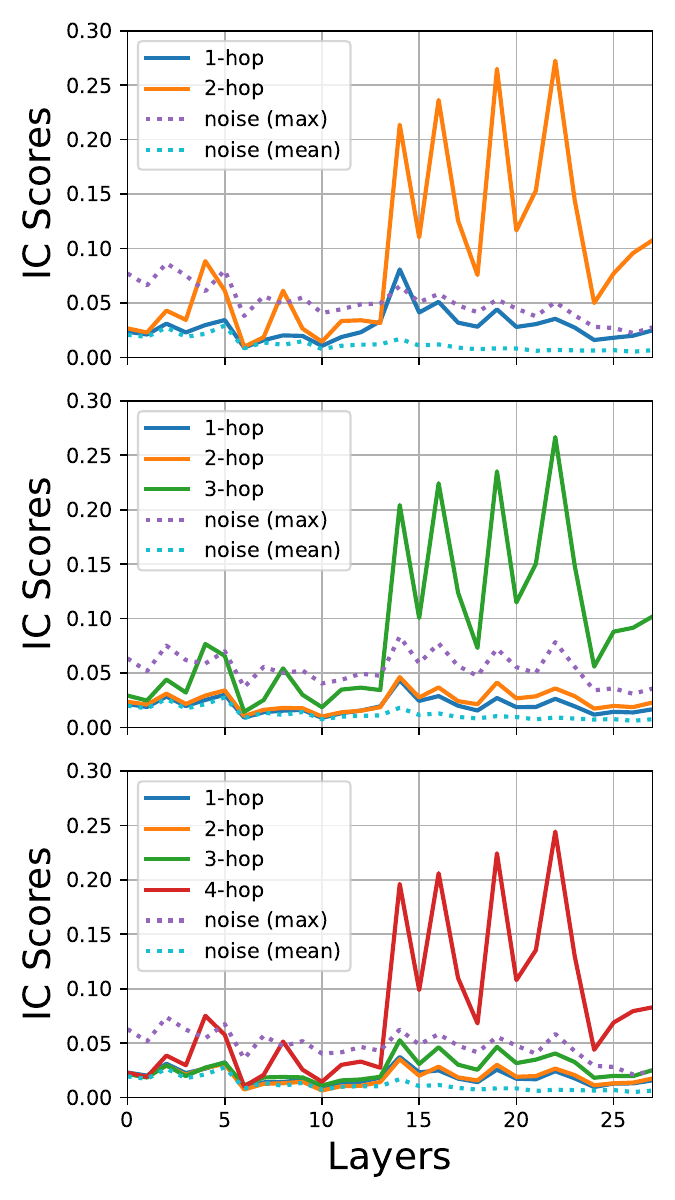}
    \caption{FT+Bi, w/ 1 hop}
    \end{subfigure} 
    \hfill
    \begin{subfigure}[b]{0.16\textwidth} 
    \centering 
    \includegraphics[width=\linewidth]{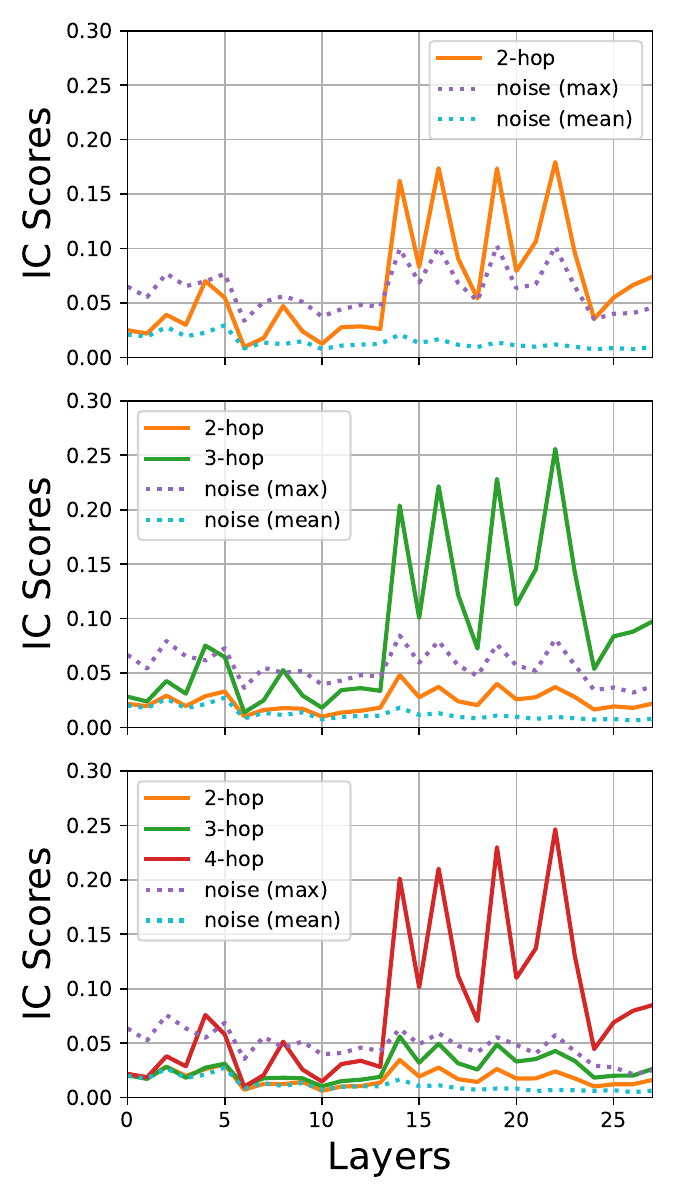}
    \caption{FT+Bi, w/o 1 hop}
    \end{subfigure} 

    \caption{IC distribution across different layers of Qwen2.5 7B with and without first hop document.}
    \label{fig:attention_layer_rm}
\end{figure*}

From Figure~\ref{fig:attention_layer_rm}, we can find that while 2-hop samples are most affected, removing the first hop document does not affect much the attention weight assigned to the last-hop documents. For 3-hop and 4-hop questions, the IC distribution of the rest documents is nearly the same as when not removing the first hop document. These findings again enhance the importance of attribution of retrieval-augmented generation, to ensure that the predicted answer is supported by the evidence from the context. In knowledge-intensive tasks, the ignorance of evidence completeness could be a critical issue that produces hallucinations.


\section{Experiment on other encoder-decoder models}
This section shows more experiment results of instruction-based encoder-decoder model. T0pp is a 11B model instruction finetuned from T5 11B with P3 datasets \citep{sanh2022multitask}, where T0 is a weaker version that trained with P3 subset. Noted that the favour of forward order disappeared. Given that P3 is a subset of Flan dataset \citep{10.5555/3618408.3619349}, we then consider that the ability is not obtained from model architecture, but from training data. Future works about which dataset from Flan triggered the favour could be established for a better understanding of the behaviour.

\begin{table}[]
    \centering
    \begin{tabular}{l|ccc}
    \toprule
        \textbf{Model} & $\Delta_{B}$ & Acc & $\Delta_{F}$ \\ \midrule
        
        T0 & 0.29 & 38.06 & -0.87 \\
        T0pp & -0.63 & 43.07 & -0.29 \\
    \bottomrule
    \end{tabular}
    \caption{Performance of other encoder-decoder models on the MuSiQue development set.}
    \label{tab:enc_dec_extra}
\end{table}

\section{Closed-book Result}
MHQA task is much more challenging than traditional QA tasks, where LMs can not well accomplish it without external knowledge. Here we accomplish a closed-book experiment setup with no context information provided to the LMs discussed in this work. Table~\ref{tab:close-book} shows the evaluation results. 
\begin{table*}[t]
    \centering
    \small

    \begin{tabular}{c|ccccc|ccccc|ccc}
    \toprule
        \multirow{2}{*}{\textbf{Model}} & \multicolumn{5}{c|}{Flan T5} & \multicolumn{5}{c|}{Qwen 2.5} & \multicolumn{3}{c}{Llama 3.x} \\
        & small & base & large & xl & xxl & 0.5B & 1.5B & 3B & 7B & 14B & 1B & 3B & 8B \\
        \midrule
        Acc & 0.54 & 1.08 & 2.19 & 3.10 & 3.97 & 2.65 & 2.48 & 5.46 & 8.32 & 10.18 & 2.77 & 5.59 & 10.88\\
    \bottomrule
    \end{tabular}
    \caption{Closed-book evaluation results on MuSiQue development set.}
    \label{tab:close-book}
\end{table*}

\section{Experiment Result on Other MHQA dataset}
\label{appendix_sec:2wiki_experiments}
To evaluate if our findings generalize to other MHQA datasets, we run extra experiments on 2WikiMultihopQA dataset \citep{ho-etal-2020-constructing}. We use 5,234 questions from the \texttt{compositional} subset and 1,549 questions from the \texttt{inference} subset, in which all the questions are 2-hop, and the context contains 10 short documents, of which 2 gold documents are provided. To investigate if finetuning generalizes to other datasets, we use the same models finetuned on MuSiQue. In this part, we evaluate LMs including Flan T5 xl, Flan T5 xxl, Qwen 2.5 7B, Llama 3.1 8B, as well as their finetuned and finetuned with bi-directional attention variants if applicable.

Table~\ref{tab:extra_em_comp} and \ref{tab:extra_em_inference} shows the evaluation results on 2WikiMultihopQA dataset. It is clear that the favour of forward document still exists in the two sets, and the use of bidirectional attention boosts performance significantly while being more robust with the order of gold documents. In addition, even though not finetuned on 2WikiMultihopQA, the two finetuned setup still obtain competitive performance. Moreover, compared to the finetuned setup with causal attention mask, the bi-directional attention get a better performance, and more robust to the order of gold documents.

\begin{table*}[t]
    \centering
    \begin{tabular}{l|ccc|ccc|ccc}
    \toprule
        \multirow{2}{*}{\textbf{Model}} & \multicolumn{3}{c|}{\textbf{Answer Only}} & \multicolumn{3}{c|}{\textbf{Finetuned}} & \multicolumn{3}{c}{\textbf{Finetuned + Bi}} \\
         & $\Delta_{B}$ & Acc & $\Delta_{F}$ 
         & $\Delta_{B}$ & Acc & $\Delta_{F}$
         & $\Delta_{B}$ & Acc & $\Delta_{F}$\\
        \midrule
        Qwen 2.5 7B & \cellcolor{red!4}{-0.44} & 42.19 & \cellcolor{green!9}{0.99} & \cellcolor{red!17}{-1.78} & 54.78 & \cellcolor{green!15}{1.55} & \cellcolor{red!2}{-0.25} & 61.04 & \cellcolor{green!7}{0.71} \\
        Llama 3.1 8B & \cellcolor{green!34}{3.40} & 51.60 & \cellcolor{red!31}{-3.19} & \cellcolor{red!9}{-0.99} & 58.98 & \cellcolor{green!4}{0.42} & \cellcolor{red!0}{-0.02} & 62.84 & \cellcolor{red!0}{-0.06} \\
        Flan T5 xl & \cellcolor{red!13}{-1.38} & 63.76 & \cellcolor{green!16}{1.60} & - & - & - & - & - & - \\
        \bottomrule
    \end{tabular}
    \caption{MHQA performance on the 2WikiMultihopQA \texttt{Compositional} development subset. $\Delta_B$ and $\Delta_F$ are performance differences between original documents and reordered (backward and forward) documents. {\color{green}{Green cells}} indicate performance improvement while {\color{red}{red cells}} indicate performance drop.}
    \label{tab:extra_em_comp}
\end{table*}

\begin{table*}[t]
    \small
    \centering
    \begin{tabular}{l|ccc|ccc|ccc|ccc}
    \toprule
        \multirow{2}{*}{\textbf{Model}} & \multicolumn{3}{c|}{\textbf{Answer Only}} & \multicolumn{3}{c|}{\textbf{CoT}} & \multicolumn{3}{c|}{\textbf{Finetuned}} & \multicolumn{3}{c}{\textbf{Finetuned + Bi}} \\
         & $\Delta_{B}$ & Acc & $\Delta_{F}$ 
         & $\Delta_{B}$ & Acc & $\Delta_{F}$
         & $\Delta_{B}$ & Acc & $\Delta_{F}$\\
        \midrule
        Qwen 2.5 7B & \cellcolor{red!14}{-1.42} & 13.82 & \cellcolor{green!23}{2.39} & \cellcolor{red!11}{-1.16} & 49.52 & \cellcolor{green!1}{0.19} & \cellcolor{red!60}{-6.00} & 33.25 & \cellcolor{green!59}{5.94} & \cellcolor{red!18}{-1.87} & 40.99 & \cellcolor{green!29}{2.97} \\
        Llama 3.1 8B & \cellcolor{red!10}{-1.03} & 24.47 & \cellcolor{red!3}{-0.39} & \cellcolor{green!5}{0.52} & 59.85 & \cellcolor{red!7}{-0.77} & \cellcolor{red!21}{-2.13} & 46.35 & \cellcolor{green!37}{3.74} & \cellcolor{red!13}{-1.36} & 56.10 & \cellcolor{green!3}{0.32}\\
        Flan T5 xl & \cellcolor{red!1}{-0.13} & 13.69 & \cellcolor{red!5}{-0.58} & - & - & - & - & - & - & - & - & - \\
        Flan T5 xxl & \cellcolor{red!31}{-3.16} & 21.11 & \cellcolor{green!25}{2.52} & - & - & - & - & - & - & - & - & - \\
        \bottomrule
    \end{tabular}
    \caption{MHQA performance on the 2WikiMultihopQA \texttt{Inference} development subset. $\Delta_B$ and $\Delta_F$ are performance differences between original documents and reordered (backward and forward) documents. {\color{green}{Green cells}} indicate performance improvement while {\color{red}{red cells}} indicate performance drop.}
    \label{tab:extra_em_inference}
\end{table*}

\begin{figure*}[t]
    \centering    
    \begin{subfigure}[b]{0.48\textwidth} 
    \centering 
    \includegraphics[width=\linewidth]{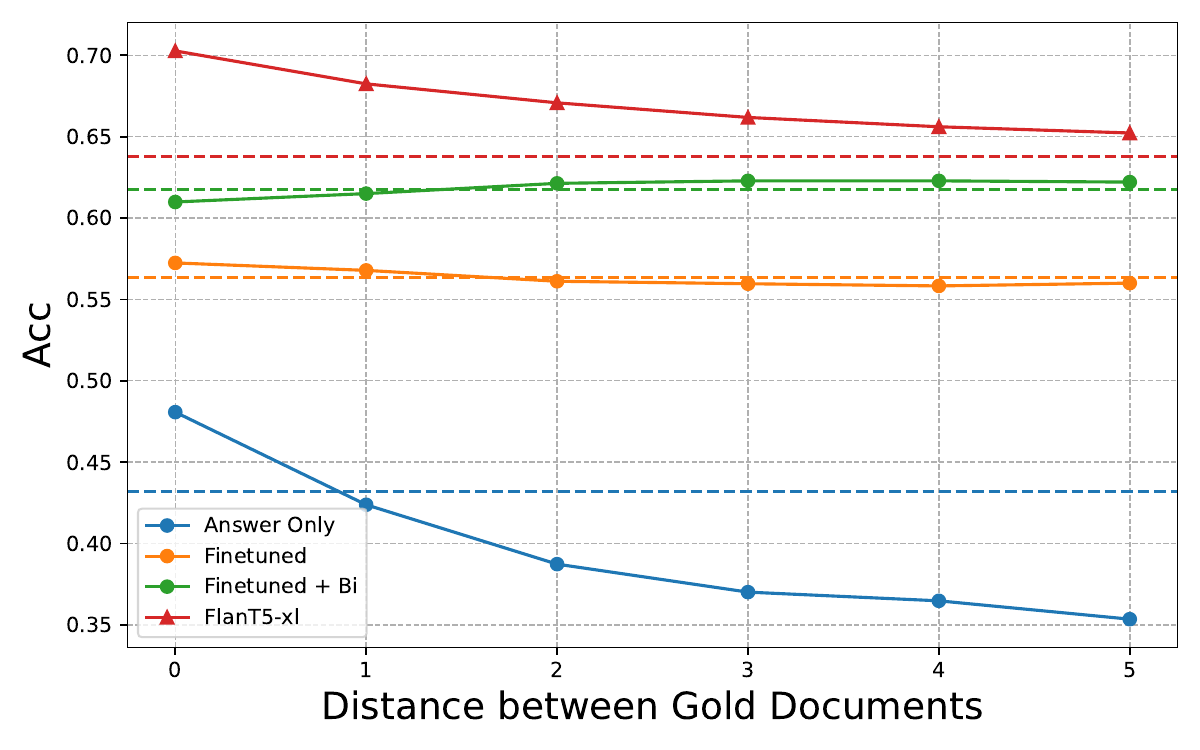}
    \caption{\texttt{Compositional}}
    \label{fig:distance_2wiki_comp}
    \end{subfigure} 
    \hfill
    \begin{subfigure}[b]{0.48\textwidth} 
    \centering 
    \includegraphics[width=\linewidth]{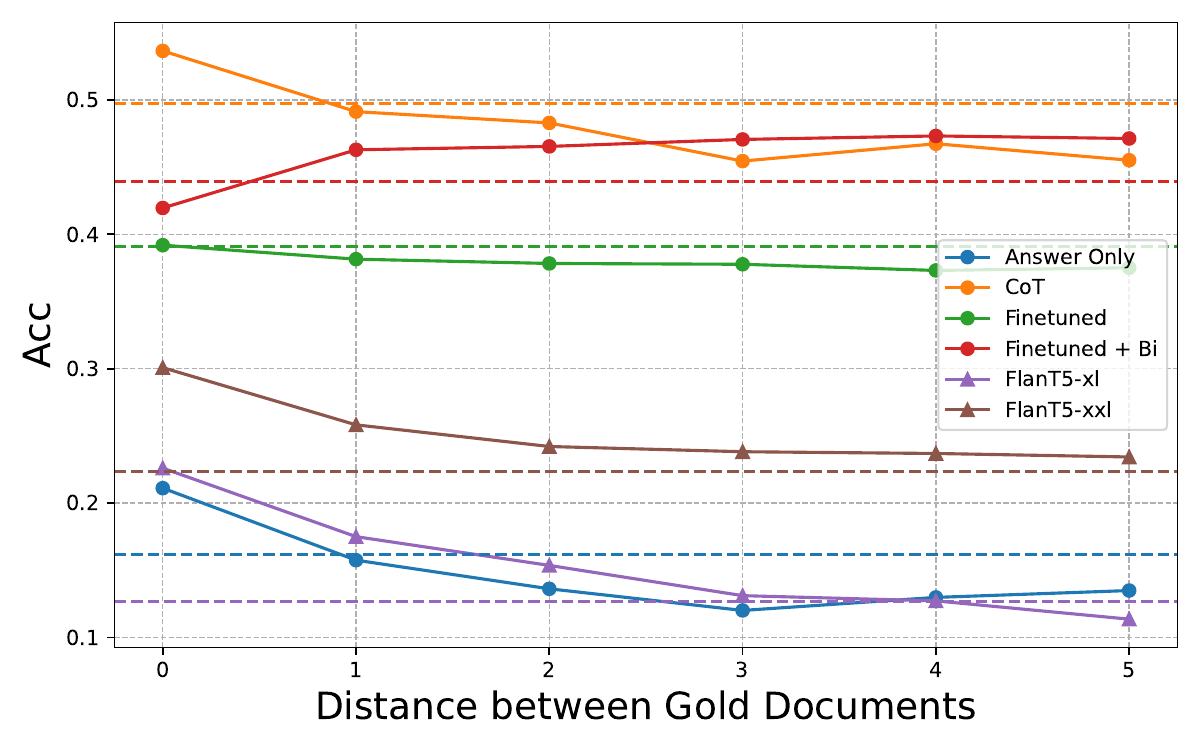}
    \caption{\texttt{Inference}}
    \label{fig:distance_2wiki_inf}
    \end{subfigure}

    \caption{Distance results on 2WikiMultihopQA development set.}
    \label{fig:distance_extra}
\end{figure*}

Figure~\ref{fig:distance_extra} shows the affect from distance on Flan T5 models and Qwen 2.5 7B variants. It is clear that our observation still holds that finetuned models are more robust to the distance of gold document, even not finetuned on the same dataset. While the performance of non-finetuned models decreases as the distance increases, even when the context is much shorter (with shorter documents and only 10 in the context).

Similar to Section~\ref{sec:peak_behave}, on the two subset of 2WikiMultihopQA dataset, we also conduct experiment on Qwen 2.5 7B with Answer Only setup with randomly shuffled the document order of each question 10 times and computed the peak IC score for each shuffle. For the \texttt{compositional} subset, the average accuracy across these 10 shuffles was 42.14, while the sample with the highest peak IC score achieved an accuracy of 49.29. While for the \texttt{inference} subset, the average accuracy across these 10 shuffles was 14.96, while the sample with the highest peak IC score achieved an accuracy of 19.43.

\end{document}